# Stacking-Based Deep Neural Network: Deep Analytic Network for Pattern Classification

Cheng-Yaw Low , Jaewoo Park, and Andrew Beng-Jin Teoh , *Senior Member, IEEE*

*Abstract*—Stacking-based deep neural network (S-DNN) is aggregated with pluralities of basic learning modules, one after another, to synthesize a deep neural network (DNN) alternative for pattern classification. Contrary to the DNNs trained from end to end by backpropagation (BP), each S-DNN layer, that is, a self-learnable module, is to be trained decisively and independently without BP intervention. In this paper, a ridge regression-based S-DNN, dubbed deep analytic network (DAN), along with its kernelization (K-DAN), are devised for multilayer feature relearning from the pre-extracted baseline features and the structured features. Our theoretical formulation demonstrates that DAN/K-DAN relearn by perturbing the intra/interclass variations, apart from diminishing the prediction errors. We scrutinize the DAN/K-DAN performance for pattern classification on datasets of varying domains—faces, handwritten digits, generic objects, to name a few. Unlike the typical BP-optimized DNNs to be trained from gigantic datasets by GPU, we reveal that DAN/K-DAN are trainable using only CPU even for small-scale training sets. Our experimental results show that DAN/K-DAN outperform the present S-DNNs and also the BP-trained DNNs, including multiplayer perceptron, deep belief network, etc., without data augmentation applied.

## I. Introduction

**D**EEP neural network (DNN) is architecturally a multilayer stack of elementary building blocks (or modules), each of which is a nonlinear interleaving layer or more sophisticatedly a subnetwork with the output of one rendering the input of the next [1]. For hierarchical representation learning (from raw inputs to high-level intricate abstractions), a nonlinear input–output mapping is learned for each in the stack in an end-to-end manner by backpropagation (BP) algorithm. To date, DNNs, in particular, multilayer perceptron (MLP)-driven instances, i.e., convolutional neural networks (CNNs, as an MLP special case) [2]–[4] and recurrent neural networks (RNNs, as a generalization for MLP) [5], [6], have accomplished significant breakthroughs for image, video, speech, and audio signals. Despite of that DNN, such as MLP, is hardly trained from the pre-extracted baseline features and the structural hand-engineered features, especially when sufficient training data is inaccessible to learn the large parameter set.

Stacking-based deep neural network (S-DNN) emerges as a DNN-alike resemblance, e.g., [10]–[13], [15], [24]–[27], [29], [30], [32]–[34], etc., for pattern classification. Architecturally, S-DNN is in line with stacked generalization [7], [8], which aggregates a chain of independent self-learnable modules. In lieu of end-to-end BP-based training, S-DNN deciphers large-scale problems via modularization, where each modular unit is engaged to learn an effective function to untangle a prefixed problem decisively and independently. Hence, there is zero, or minimal interaction between any two neighboring modules. In principal, there is no restriction applied to the layer-wise learner selection, as long as a meaningful mapping is realized. The broadly adopted learners are principal component analysis (PCA) [9], linear discriminant analysis (LDA) [14], ridge regression (RR) [20], extreme learning machine (ELM) [28], and random forest (RF) [31]. In general, S-DNNs are either be convolutional or nonconvolutional. The convolutional S-DNN receives only images for feature extraction via image-filter convolutions, followed by an optional feature encoding stage. Some of the pertinent works are the PCA network (PCANet) and its variants [10]–[13], and the LDA-learned deep discriminant face descriptor (D-DFD) [15]. Different from that of convolutional, the nonconvolutional S-DNN is topologically fully connected (FC) for both images and nonimages such as the pre-extracted baseline features and also the handcrafted structured features. The most representative FC S-DNNs include deep convex networks (DCNs) [24]–[27]; deep ELMs (D-ELMs) [29], [30]; and deep forests (DFs) [32]–[34]. We group all the abovementioned networks under the S-DNN umbrella term in Fig. 1.

On the other hand, there are relevant S-DNNs relying on BP for global fine-tuning (FT) on the modularly trained networks, e.g., deep belief networks (DBNs) [16], [17]; deep Boltzmann machine (DBM) [18]; and deep auto-encoder (DAE) [19]. We refer these exceptional S-DNN to S-DNN with BP (S-DNN$_{BP}$) for performance analysis and comparison. This paper, however, focuses only on S-DNN trained conveniently without BP.

Manuscript received August 28, 2018; revised November 17, 2018; accepted March 26, 2019. This work was supported in part by the National Research Foundation of Korea (NRF) grant funded by the Korea Government (MSIP) under Grant NRF-2019R1A2C1003306. This paper was recommended by Associate Editor Y. Zhang. *(Corresponding author: Andrew Beng-Jin Teoh.)*
C.-Y. Low was with the School of Electrical and Electronic Engineering, College of Engineering, Yonsei University, Seoul 03722, South Korea. He is now with the Faculty of Information Science and Technology, Multimedia University, Melaka 75450, Malaysia (e-mail: cylow@mmu.edu.my).

J. Park and A. B.-J. Teoh are with the School of Electrical and Electronic Engineering, College of Engineering, Yonsei University, Seoul 03722, South Korea (e-mail: julypraise@yonsei.ac.kr; bjteoh@yonsei.ac.kr).

Color versions of one or more of the figures in this paper are available online at http://ieeexplore.ieee.org.

Digital Object Identifier 10.1109/TCYB.2019.2908387

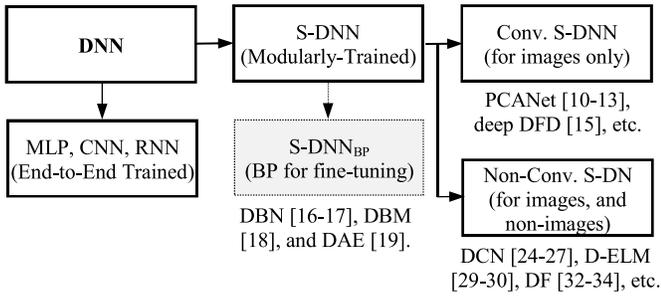

Fig. 1. DNN and subdisciplines.

*A. Related Works*

This paper underlines the FC S-DNN trained with no BP involvement. One of the earliest S-DNN is the supervised DCN [24]. Each DCN layer is a single-hidden-layer network with the sigmoidal activations yielding a regression output set to the immediate next layer. The DCN innovation is afterward extended to the kernelized-DCN (K-DCN) [25], [26] and the tensor deep stacking network (T-DSN) [27]. These DCN instances are disclosed to be on par with the BP-tuned DBN [17].

The D-ELM, e.g., stacked ELM (S-ELM) [29], autoencoder-based S-ELM (AE-S-ELM) [29], hierarchical ELM (H-ELM) [30], etc., outline the ELM-based S-DNNs. To address the computational burden for high-dimensional random projection, each ELM module for S-ELM and AE-S-ELM is designated with a streamlined hidden node set in exchange for deeper construction. On the other hand, both AE-S-ELM and H-ELM train the ELM-based autoencoder for the refined random input-hidden weights. Overall, these D-ELMs outshine the traditional single-layer ELM, support vector machine (SVM), and DBNs.

The most recent S-DNN is the DF by Zhou and Feng [32], where each layer is an ensemble of random decision tree forests, or an ensemble of ensembles. To assure high degree of diversity leading to performance gain in consequence, each ensemble is instantiated with RFs of different types. Other DF-driven S-DNNs include [33] and [34]. However, the performance evaluation of these DF networks is limited to only handwritten digit recognition.

*B. Motivation*

The BP-optimized DNNs, e.g., CNNs and RNNs, have been demonstrated prominent in learning representative hierarchical features from image, video, speech, and audio inputs. However, to the best of our knowledge, there is no DNN learnable on top of the pre-extracted rudimentary features elicited from images, e.g., [10], [42], and [43], and the nonsignal/nonsequential data [41], e.g., lab measurements, social-demographic variables, and human annotated examples of which we term as the handcrafted structured features in this paper. One possible alternative is the MLP empowered with the layer-wise pretraining [16], [17]. The cornerstones of training an arbitrary deep and high-complexity MLP are: 1) MLP only relies on the iterative BP algorithm; 2) a gigantic training set is demanded to confront overfitting issue; 3) network training is a black box due to the lack of theoretical grounds defined to fine-tune the massive hyper-parameter set; 4) network adaptability and scalability problems, e.g., any amendments to a pretrained MLP requires retraining from the scratch; and 5) GPU employment is of mandatory for training.

In comparison to MLP (and other BP-trained DNNs), the S-DNN with no BP intervention stands out in four perspectives: 1) fast learning speed owing to no BP and no mysterious hyper-parameter tuning; 2) no gigantic data demanded as training is of module-based, one after another; 3) modularly stacked networks are adaptable and scalable; and 4) training applies no GPU but only CPU as complexity is reasonably inexpensive.

*C. Contribution*

This paper is inspired by the hierarchical representation learning in DNNs. The three contributions are as follows.
1) An RR-based S-DNN, i.e., deep analytic network (DAN) and its kernelization (K-DAN), are outlined to learn a non-BP S-DNN involving no GPU, no enormous training set, and no elusive hyper-parameter tuning.
2) DAN/K-DAN are attested triggering feature relearning from the pre-extracted baseline features and the structured features, of which CNNs and RNNs are impracticable. Under a certain condition that the relearned feature dimension is outnumbered by that of original, DAN/K-DAN perform also feature compression. This is overlooked and thus not being explored thoroughly in other relevant works.
3) DAN/K-DAN are analyzed for proofs contributing to the improved generalizability. We portray the basic self-learnable unit to assemble the deep DAN/K-DAN construction in Fig. 2, and the complete DAN/K-DAN pipeline is illustrated in Fig. 3.

This paper is an extension to our preliminary work that only emphasizes DAN [44]. With the twofold PCA filter-to-filter convolution features (2-FFC$_{PCA}$) [43], we summarize in that paper the extent to which DAN improves the 2-FFC$_{PCA}$ baseline performance without any theoretical justifications. For further analysis and exploration, we also outline K-DAN in this paper on top of DAN. We conduct extensive experiments to examine the DAN/K-DAN aptitude for feature relearning, including relearning from the pre-extracted DNN features. For comparison to that of DNNs in terms of most primitive performance, computational complexity, and CPU training and interference time, we retrain MLP [53] and other influential BP-optimized DNNs [54]–[57] with only a single network without applying data augmentation. We demonstrate in Section V that DAN/K-DAN outperform the BP-trained counterparts, aside from being the most promising among the other S-DNNs. Our implementation codes are available on GitHub for result replication as follows: https://github.com/chengyawlow/DAN.

*D. Organization*

The organization of this paper is deliberated as follows: RR/KRR, as the single-module equivalences for DAN/K-DAN, are formulated in Section II, followed by the algorithmic details in Section III. The supporting theories are elucidated in Section IV, and we summarize in Section V the empirical performance for DAN/K-DAN and other counterparts. Subsequent to that a concluding remark, along with future works, are provided in the last section.

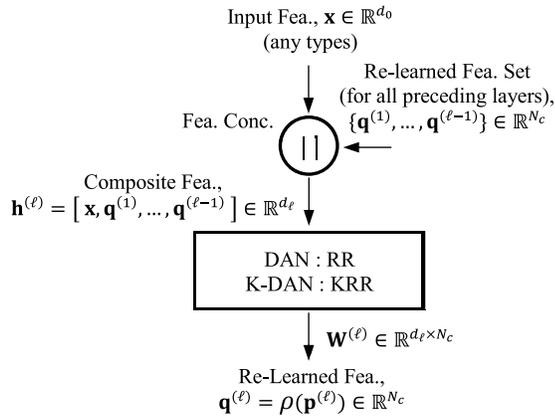

Fig. 2. Self-learnable building block for DAN and K-DAN.

## II. PRELIMINARIES

In principal, DAN/K-DAN are of parallel to that of multilayer (deep) RR/KRR [20] constructions interleaved with the rectified linear unit (ReLU) [51] for feature relearning. We thus delineate RR/KRR as the groundworks for DAN/K-DAN in this section.

### A. Ridge Regression

Suppose $\{(\mathbf{x}_i, \mathbf{y}_i)\}_{i=1}^N$ be a set of $N$ training samples; each $\mathbf{x}_i \in \mathbb{R}^d$ is associated with a target vector $\mathbf{y}_i \in \{\mathbf{t}_1, \mathbf{t}_2, \ldots, \mathbf{t}_{N_c}\}$, where $\mathbf{t}_j = [t_{j,1}, \ldots, t_{j,j}, \ldots, t_{j,N_c}] \in \mathbb{R}^{N_c}$ is of one-hot encoded with the only $j$th element (conforming to the class label) set to one whereas the remaining are of zeros, and $N_c$ represents the number of training classes. Let $\mathbf{X} = [\mathbf{x}_1, \ldots, \mathbf{x}_N]^T \in \mathbb{R}^{N \times d}$ and $\mathbf{Y} = [\mathbf{y}_1, \ldots, \mathbf{y}_N]^T \in \mathbb{R}^{N \times N_c}$; to resort the ill-posed least squares formulation from singularity problem, RR learns the regression coefficients, i.e., the weight matrix $\mathbf{W} \in \mathbb{R}^{d \times N_c}$, by minimizing the penalized sum of squares $L(\mathbf{W})$ as follows:

$$L(\mathbf{W}) = \text{tr}\left[(\mathbf{Y} - \mathbf{X}\mathbf{W})^T(\mathbf{Y} - \mathbf{X}\mathbf{W})\right] + \lambda \|\mathbf{W}\|_F^2 \quad (1)$$

where $\lambda$ is an RR regularization parameter, and $\|.\|_F$ denotes the Frobenius norm. Assuming that $N \geq d$, $\mathbf{W}$ is estimated as follows:

$$\mathbf{W} = \left(\mathbf{X}^T\mathbf{X} + \lambda \mathbf{I}\right)^{-1} \mathbf{X}^T \mathbf{Y} \quad (2)$$

where $\mathbf{I}$ denotes an identity matrix of relevant dimension. On the other hand, for $N < d$, (2) is rewritten into its equivalence as in

$$\mathbf{W} = \mathbf{X}^T\left(\mathbf{X}\mathbf{X}^T + \lambda \mathbf{I}\right)^{-1} \mathbf{Y}. \quad (3)$$

For an unknown sample $\mathbf{x}_{te} \in \mathbb{R}^d$, the pretrained RR returns a response vector $\hat{\mathbf{y}}_{te} \in \mathbb{R}^{N_c}$ as follows:

$$\hat{\mathbf{y}}_{te} = \mathbf{W}^T \mathbf{x}_{te}. \quad (4)$$

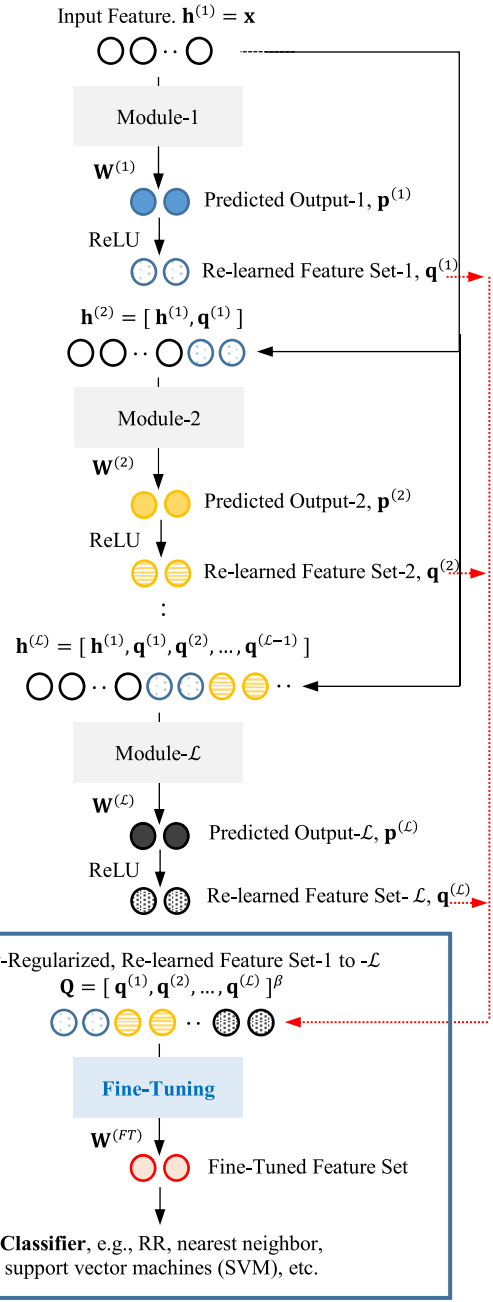

Fig. 3. Generic DAN construction of $\mathcal{L}$ self-learnable layers affixed with an FT module (bounded in blue). Each layer $\ell$ yields a relearned feature set $\mathbf{q}^{(\ell)}$ from $\mathbf{h}[]^{(\ell)}$, and the relearned feature sets for all $\mathcal{L}$ layers are aggregated into power-regularized feature set $\mathbf{Q}$ for FT to learn an auxiliary classifier.

Subsequent to that, the label of $\mathbf{x}_{te}$ can be determined based on $\hat{\mathbf{y}}_t$ as in

$$cls(\mathbf{x}_{te}) = \underset{j \in \{1, \ldots, N_c\}}{\text{argmax}} \left(\hat{\mathbf{y}}_{te}\right)_j. \quad (5)$$

### B. Kernel Ridge Regression

Kernel machine, i.e., a machine learning model employing a prespecific nonlinear function via a neat kernel trick, appeared in 1960s [21]. This emergence contributed to the earliest kernel machine [22], followed by the great accomplishments of kernel SVMs in 1990s [23]. Other linear models that which can be kernelized are RR, PCA, LDA, etc.

KRR and other kernel machines operate in an implicit feature space of (infinitely) high dimension. In other words, rather than an explicit transformation, only the inner products over training sample pairs are measured based on an arbitrary positive semidefinite kernel function $k$ that satisfies the Mercer's condition. The well-known kernel functions are the polynomial kernel, the radial basis function (RBF) kernel, the Laplacian kernel, etc.

To analytically estimate $\mathbf{W}$, KRR recasts (3) as follows:

$$\mathbf{W} = \mathbf{X}^T(\mathbf{K} + \lambda \mathbf{I})^{-1}\mathbf{Y} \tag{6}$$

where $\mathbf{K} \in \mathbb{R}^{N \times N}$ is the Gram matrix (or kernel matrix) with $\mathbf{K}_{ij} = k(\mathbf{x}_i, \mathbf{x}_j)$. In accordance to (6), KRR yields $\hat{\mathbf{y}}_{te}$ for $\mathbf{x}_{te}$ as follows:

$$\hat{\mathbf{y}}_{te} = \mathbf{Y}^T(\mathbf{K} + \lambda \mathbf{I})^{-1} \begin{bmatrix} k(\mathbf{x}_{te}, \mathbf{x}_1) \\ \vdots \\ k(\mathbf{x}_{te}, \mathbf{x}_N) \end{bmatrix}^T. \tag{7}$$

Along with that the corresponding class label is predicted to be $j$ as in (5).

## III. Deep Analytic Network and Kernelization

Similar to that of DCNs and D-ELMs reviewed in the foregoing sections, we assemble DAN and K-DAN upon RR and kernel RR (KRR) [20], respectively. The main reason is that in place of the iterative BP-trained gradient descent algorithm, RR provides an analytic solution leading the learning progression parsimonious. This section explicates how a series of RR/KRR-based building blocks, being the basic self-learnable module each, are stacked into the deep, feedforward DAN/K-DAN.

### A. Basic Self-Learnable Module

As shown in Fig. 2, the internal constructions for each DAN/K-DAN building block are different in that one is of RR-based and another operates on KRR. By cascading multiple self-learnable modules, the $\ell$th layer of DAN/K-DAN learns based on $\mathbf{h}^{(\ell)} = [\mathbf{x}, \mathbf{q}^{(1)}, \dots, \mathbf{q}^{(\ell-1)}] \in \mathbb{R}^{d_\ell}$, i.e., a stacking vector comprising of the input feature $\mathbf{x}$ of any types and other relearned features for all preceding modules $\{\mathbf{q}^{(i)}\}_1^{\ell-1}$ with dimension $d_\ell = d + N_c(\ell - 1)$ to yield a new relearned feature set $\mathbf{q}^{(\ell)}$. Considering that each modular unit, regardless of RR or KRR, responds with $\mathbf{p}^{(\ell)} \in \mathbb{R}^{N_c}$ to be nonlinearly projected into $\mathbf{q}^{(\ell)} \in \mathbb{R}^{N_c}$ such that $\mathbf{q}^{(\ell)} = \rho(\mathbf{p}^{(\ell)})$ In this paper, $\rho(.)$ denotes the ReLU activation function performing $\mathbf{q}^{(\ell)} = \max(0, \mathbf{p}^{(\ell)})$. To navigate into a deeper construction of $\ell + 1$, $\mathbf{h}^{(\ell+1)} \in \mathbb{R}^{d_{\ell+1}}$ composing of $\mathbf{x}$ and the stack of $\mathbf{q}^{(1)}, \mathbf{q}^{(2)}, \dots, \mathbf{q}^{(\ell)}$ is formed to yield $\mathbf{q}^{(\ell+1)}$ of generally more discriminative. Provided with the training repository $\mathbf{X}$ with $N$ samples; we transform $\mathbf{P}^{(\ell)} = [\mathbf{p}_1^{(\ell)}, \dots, \mathbf{p}_N^{(\ell)}]^T \in \mathbb{R}^{N \times N_c}$ into $\mathbf{Q}^{(\ell)} = [\mathbf{q}_1^{(\ell)}, \dots, \mathbf{q}_N^{(\ell)}]^T \in \mathbb{R}^{N \times N_c}$. Along with that we derive $\mathbf{H}^{(\ell)} = [\mathbf{h}_1^{(\ell)}, \dots, \mathbf{h}_N^{(\ell)}]^T \in \mathbb{R}^{N \times d_\ell}$ accordingly to estimate $\mathbf{W}^{(\ell)} \in \mathbb{R}^{d_\ell \times N_c}$ based on RR, or KRR (refer to Sections III-B and III-C).

Unlike DAN triggering feature relearning on $\mathbf{h}^{(\ell)}$ directly, each KRR modular unit in K-DAN performs RR in the implicit nonlinear transformed space. In other words, a DAN module is an FC layer implementing the nonlinear RR-trained building block; while each K-DAN module is a specialized two-layer network exercising kernelization prior to the RR feature relearning stage. In a nutshell, every single DAN/K-DAN layer nonlinearly maps $\mathbf{h}^{(\ell+1)}$ to the relearned feature set $\mathbf{q}^{(\ell+1)}$ for further exploration in all the succeeding layers. The complete DAN/K-DAN construction is detailed in the following sections.

### B. Deep Analytic Network

The DAN construction is stacked with the RR-learned units for layer-wise feature relearning, as portrayed in Fig. 3. Following the definitions in Section II, assuming $N \geq d$, the first layer of DAN with depth $\mathcal{L}$ is delivered with $\mathbf{X}$, and (2) is extended to estimate the analytic weight set $\mathbf{W}^{(\ell)} \in \mathbb{R}^{d_\ell \times N_c}$ for $\ell = 1, \dots, \mathcal{L}$ as follows:

$$\mathbf{W}^{(\ell)} = \left(\mathbf{H}^{(\ell)T}\mathbf{H}^{(\ell)} + \lambda^{(\ell)}\mathbf{I}\right)^{-1}\mathbf{H}^{(\ell)T}\mathbf{Y} \tag{8}$$

where $\mathbf{H}^{(1)} = \mathbf{X}$, $\mathbf{H}^{(2 \leq \ell \leq \mathcal{L})} = [\mathbf{H}^{(1)}, \mathbf{Q}^{(1)}, \dots, \mathbf{Q}^{(\ell-1)}] \in \mathbb{R}^{N \times d_\ell}$ or equivalently $\mathbf{H}^{(2 \leq \ell \leq \mathcal{L})} = [\mathbf{H}^{(\ell-1)}, \mathbf{Q}^{(\ell-1)}]$. Note in (9) that $\mathbf{Q}^{(\ell)}$ accommodates the ReLU-transformed response matrix of $\mathbf{P}^{(\ell)} = \mathbf{H}^{(\ell)}\mathbf{W}^{(\ell)}$, such that $\mathbf{P}^{(\ell)}, \mathbf{Q}^{(\ell)} \in \mathbb{R}^{N \times N_c}$. The supporting theories in Section IV underscores that this non-negativity is of crucial in two perspectives: 1) the ReLU-activated responses improve the training prediction accuracy and 2) DAN forms the dynamics of inter/intraclass distances; we solidify these theories based on the empirical observations in Section V

$$\begin{aligned} \mathbf{P}^{(\ell)} &= \mathbf{H}^{(\ell)}\mathbf{W}^{(\ell)} \\ \mathbf{Q}^{(\ell)} &= \rho\left(\mathbf{P}^{(\ell)}\right) = \max\left(0, \mathbf{P}^{(\ell)}\right). \end{aligned} \tag{9}$$

The DAN FT output layer with the power-law nonlinearity is the last building block on the stack, which embeds the built-in regression classifier. We power-regularize all the relearned feature sets $\{\mathbf{Q}^{(\ell)}\}_1^\mathcal{L}$ with respect to a small positive ratio of $0 \leq \beta \leq 1$ in the element-wise manner to synthesize $\mathbf{Q}_{FT} = [\mathbf{Q}^{(1)}, \dots, \mathbf{Q}^{(\mathcal{L})}]^\beta \in \mathbb{R}^{N \times d_{FT}}$, where $d_{FT} = \mathcal{L} \times N_c$. This is to regularize the disparities within the relearned features before $\mathbf{W}_{FT} \in \mathbb{R}^{d_{FT} \times N_c}$ is learned as follows:

$$\mathbf{W}_{FT} = \left(\mathbf{Q}_{FT}^T\mathbf{Q}_{FT} + \lambda_{FT}^{(\ell)}\mathbf{I}\right)^{-1}\mathbf{Q}_{FT}^T\mathbf{Y}. \tag{10}$$

Depending on the task at hand, the FT layer is opted for other classifiers, e.g., nearest neighbor (NN) classifier, SVMs, etc.

Regardless of the classifier types, the response vector $\hat{\mathbf{y}}_{te}$ for $\mathbf{x}_{te}$ is estimated as follows:

$$\hat{\mathbf{y}}_{te} = \mathbf{W}_{FT}^T \mathbf{q}_{FT, te} \tag{11}$$

where $\mathbf{q}_{FT, te} = [\mathbf{q}_{te}^{(1)}, \dots, \mathbf{q}_{te}^{(\mathcal{L})}]^\beta \in \mathbb{R}^{d_{FT}}$. Our formulation derives $\mathbf{q}_{te}^{(\ell)} = \max(0, \mathbf{p}_{te}^{(\ell)})$, where $\mathbf{p}_{te}^{(\ell)} = \mathbf{W}^{(\ell)T}\mathbf{h}_{te}^{(\ell)}$ and $\mathbf{h}_{te}^{(\ell)} = (x_{te}, \mathbf{q}_{te}^{(1)}, \dots, \mathbf{q}_{te}^{(\ell-1)}) \in \mathbb{R}^{D_\ell}$. Considering the default classifier in (5), the class label for $\mathbf{x}_{te}$ is inferred as $j$.

In summary, the DAN feature relearning stage involves a set of three parameters to be fine-tuned, specifically $\lambda^{(\ell)}$, $\lambda_{FT}^{(\ell)}$, and $\beta_{FT}$. Our experiments dispatch DAN with the pre-extracted 2-FFC$_{PCA}$ features [43] for multilayer feature relearning, unless specified otherwise. To be more precise, the DAN relearning procedures are summarized in Table I.

TABLE I
DAN FEATURE RELEARNING PROGRESSION

| DAN |
|---|
| The DAN inputs are:<br>  i. Pre-extracted baseline features, $\mathbf{X} \in \mathbb{R}^{N \times d}$;<br>  ii. One-hot encoded, $\mathbf{Y} = [\mathbf{y}_1, \ldots, \mathbf{y}_N]^T \in \mathbb{R}^{N \times N_c}$;<br>  iii. Regression parameters: $\lambda^{(\ell)}$ and $\lambda_{FT}^{(\ell)}$.<br>  iv. Power regularization ratio $\beta_{FT}$. |
| Step 1 : If $\ell = 1$, $\mathbf{H}^{(1)} = \mathbf{X}$.<br>Otherwise, $\mathbf{H}^{(\ell)} = [\mathbf{H}^{(1)}, \mathbf{Q}^{(1)}, \ldots, \mathbf{Q}^{(\ell-1)}] \in \mathbb{R}^{N \times d_\ell}$,<br>where $d_\ell = d + N_c(\ell - 1)$. |
| Step 2 : Compute $\mathbf{W}^{(\ell)} \in \mathbb{R}^{d_\ell \times N_c}$ with respect to (8). |
| Step 3 : Compute $\mathbf{Q}^{(\ell)} = \max(0, \mathbf{P}^{(\ell)})$, where $\mathbf{P}^{(\ell)} = \mathbf{H}^{(\ell)}\mathbf{W}^{(\ell)}$, and $\mathbf{P}^{(\ell)}, \mathbf{Q}^{(\ell)} \in \mathbb{R}^{N \times N_c}$. |
| Step 4 : If $\ell \neq \mathcal{L}$, repeat Step 1 to Step 3 until $\ell = \mathcal{L}$. |
| Step 4.1 : Otherwise, $\mathbf{Q}_{FT} = [\mathbf{Q}^{(1)}, \ldots, \mathbf{Q}^{(\mathcal{L})}]^\beta$,<br>where $\mathbf{Q}_{FT} \in \mathbb{R}^{N \times d_{FT}}$, and $d_{FT} = \mathcal{L} \times N_c$. |
| Step 4.2 : Compute $\mathbf{W}_{FT}^{(\ell)} \in \mathbb{R}^{d_{FT} \times N_c}$ with respect to (10). |

TABLE II
K-DAN FEATURE RELEARNING PROGRESSION

| K-DAN |
|---|
| The K-DAN inputs are:<br>  i. Raw input features, e.g., image, or non-image features, including the structural hand-engineered features, $\mathbf{X} \in \mathbb{R}^{N \times d}$;<br>  ii. One-hot encoded, $\mathbf{Y} = [\mathbf{y}_1, \ldots, \mathbf{y}_N]^T \in \mathbb{R}^{N \times N_c}$;<br>  iii. Regression parameters: $\lambda^{(\ell)}$, and $\lambda_{FT}^{(\ell)}$.<br>  iv. RBF-kernel spreading factor $\gamma^{(\ell)}$.<br>  v. Power regularization ratio $\beta_{FT}$.<br>The K-DAN feature re-learning progression follows DAN, except in **Step 2**, where $\mathbf{W}^{(\ell)} \in \mathbb{R}^{d_\ell \times N_c}$ is estimated with respect to (12). |

### C. Kernel Deep Analytic Network

We consider only K-DAN with the RBF kernel in this paper, albeit other Mercer kernels are possible. Pursuant to the KRR formulation in Section II, K-DAN rewrites (6) as follows:

$$\mathbf{W}^{(\ell)} = \mathbf{H}^{(\ell)T}\left(\mathbf{K}^{(\ell)} + \lambda^{(\ell)}\mathbf{I}\right)^{-1}\mathbf{Y} \quad (12)$$

where $\mathbf{K}^{(\ell)}$ denotes the RBF-defined kernel matrix with $\mathbf{K}_{i,j}^{(\ell)} = k(\mathbf{h}_i^{(\ell)}, \mathbf{h}_j^{(\ell)}) = \exp(-\gamma^{(\ell)}\|\mathbf{h}_i^{(\ell)} - \mathbf{h}_j^{(\ell)}\|^2)$, such that $\gamma^{(\ell)}$ is an empirical parameters controlling bias and variance. As in DAN, the K-DAN relearning phase demands also $\mathbf{P}^{(\ell)}$ and $\mathbf{Q}^{(\ell)}$ to be described as in (9) to elaborate $\mathbf{H}^{(\ell)}$. Subsequent to that, $\mathbf{W}_{FT}$ is learned from $\mathbf{Q}_{FT}$ as in (10).

To yield $\mathbf{q}_{te}^{(\ell)}$ from for $\mathbf{x}_{te}$ where $\mathbf{q}_{te}^{(\ell)} = \max(0, \mathbf{p}_{te}^{(\ell)})$, (7) in KRR is revised as follows:

$$\mathbf{p}_{te}^{(\ell)} = \mathbf{Y}^T\left(\mathbf{K}^{(\ell)} + \lambda^{(\ell)}\mathbf{I}\right)^{-1}\begin{bmatrix} k\left(\mathbf{h}_{te}^{(\ell)}, \mathbf{H}_1^{(\ell)}\right) \\ \vdots \\ k\left(\mathbf{h}_{te}^{(\ell)}, \mathbf{H}_N^{(\ell)}\right) \end{bmatrix}^T. \quad (13)$$

Following that $\hat{\mathbf{y}}_{te}$ is elicited based on the prelearned $\mathbf{W}_{FT}$ and $\mathbf{q}_{FT,te}$ as in (11) to predict the class label for $\mathbf{x}_{te}$ by the regression classifier in (5). As a whole, K-DAN encapsulates four empirical parameters: $\lambda^{(\ell)}$, $\gamma^{(\ell)}$, $\lambda_{FT}^{(\ell)}$, and $\beta_{FT}$. We provide the K-DAN relearning summary in Table II.

### D. Comparison to DNNs and Existing S-DNNs

In general, DAN/K-DAN exhibit the four S-DNN attributes outlined in Section I-B. These include the fast learning advantage resulted from the RR/KRR employment for one-shot (analytic) solution. Training DAN/K-DAN hence requires only minimal efforts owing to no massive training data, no BP, and no GPU. Moreover, the modularly trained DAN/K-DAN removes the network depth $\mathcal{L}$ from the hyper-parameter list. A new module is conveniently introduced to the existing stack based on the layer-wise performance of the validation set. In addition to that opposing to DNNs, the number of hidden nodes for each layer $d_\ell$ is of deterministic such that $d_\ell = d + N_c(\ell - 1)$ for $\ell = 1, \ldots, \mathcal{L}$. As DAN possesses only three hyper-parameters (and four for K-DAN), hyper-parameter tuning is of nontrivial essentially.

DAN/K-DAN simplify the existing S-DNNs, including DCN [24]–[27] and D-ELM [29]–[30]. One of the distinguishable traits is that every DAN module is of single-layer implementing RR (or KRR for K-DAN). The sigmoidal input-hidden projection based on the stochastic weights (as in S-ELM [29]), or other iteratively learned weights, either RBM-learned (as in DCN [24]), gradient descent-learned (as in T-DSN [27]), or autoencoder-learned (as in AE-S-ELM [29] and H-ELM [30]), are of nonexistent. Being the finest DCN variant learning third-order bilinear mappings, T-DSN also appears to be far-fetched. There is no significant improvement witnessed, despite of being computationally more expensive than that of DCN. On the other hand, S-ELM and AE-S-ELM, that is, the best-performing among D-ELMs, are reported requiring very deep network construction (to be exposed in Section V). More importantly, there is no theoretical analysis derived, but some black-box performance summaries evaluated on the toy datasets.

In comparison to DF assembled upon random forests [31], DAN/K-DAN consider no additional performance factors, except the regression shrinkage and the kernel regularization parameters. For DF, to ensure the diversity for each ensemble, the internal parameters to be accounted for are forest types in each ensemble, the number of forests, the number of decision trees in each forest, the tree growth, etc. We summarize in Section V that the DAN trained based on the pre-extracted basic features remarkably outperforms DF [32].

### IV. SUPPORTING THEORIES

This section formulates the DAN/K-DAN supporting theories. In particular, we disclose how the layer-wise relearned features (yielded by applying ReLU to the regression prediction) reduce the training prediction errors. Following that we analyze how the intra/interclass distances at each layer of DAN/K-DAN are changed as the layer deepens.

### A. Relearned Features by ReLU Improves Training Accuracy

Following our previous definitions, we pair each $\mathbf{x}_i$ of $\mathbf{X}$ with a binary label $y_i$ forming $\mathbf{y} = [y_1, \ldots, y_N]^T$. The prediction vector $\hat{\mathbf{y}} = \mathbf{XW}$ elicits the next layer of DAN, $\bar{\mathbf{X}} = [\mathbf{X}, \rho(\hat{\mathbf{y}})]$, with its prediction $\hat{\bar{\mathbf{y}}} = \bar{\mathbf{X}} \bar{\mathbf{W}}$, where $\rho(\hat{\mathbf{y}})$ is the relearned features yielded by applying ReLU activation to $\hat{\mathbf{y}}$, provided that $\mathbf{W}$ and $\bar{\mathbf{W}}$ be the RR projection matrices learned from $\mathbf{X}$ and $\bar{\mathbf{X}}$, respectively.

*Definition:* Let $\mathbf{z} = [z_1, \ldots, z_N] \in \mathbb{R}^N$ be a vector. Define $I(\mathbf{z} < 0) := \{i \in [N] : z_i < 0\}$ and $I(\mathbf{z} \geq 0) := \{i \in [N] : z_i \geq 0\}$ where $[N] = \{1, \ldots, N\}$. For an index set $I \subseteq [N]$, let $I^c := \{i \in [N] : i \notin I\}$ and $\mathbf{Q}_I$ be a matrix such that $\mathbf{Q}_I \mathbf{z} \in \mathbb{R}^N$ with $(\mathbf{Q}_I \mathbf{z})_i = z_i$ if $i \in I$ and 0, otherwise.

*Theorem 1:* Suppose $\dim(\mathrm{span}(\mathbf{X})) < N$. If either $(\hat{\mathbf{y}} - \mathbf{y})^T \mathbf{Q}_I \hat{\mathbf{y}} \neq 0$, or $(\hat{\mathbf{y}} - \mathbf{y})^T \mathbf{Q}_{I^c} \hat{\mathbf{y}} \neq 0$, then

$$\rho(\hat{\mathbf{y}}) \notin \mathrm{span}(\mathbf{X}) \tag{14}$$

where $I = I(\hat{\mathbf{y}} < 0)$. In this case,

$$\mathrm{span}(\mathbf{X}) < \mathrm{span}(\bar{\mathbf{X}}) \tag{15}$$

and therefore,

$$\hat{\mathbf{y}} - \mathbf{y} \geq \hat{\bar{\mathbf{y}}} - \mathbf{y}. \tag{16}$$

This asserts that if either of the $\mathbf{Q}_I$ and $\mathbf{Q}_{I^c}$ projections of the prediction samples $\hat{\mathbf{y}}$ is *not orthogonal* to the errors of the samples $\hat{\mathbf{y}} - \mathbf{y}$, the ReLU-ed prediction $\rho(\hat{\mathbf{y}})$ is outside the span of the data matrix $\mathbf{X}$. In which case, the span of $\bar{\mathbf{X}}$ gains an extra dimension by $\rho(\hat{\mathbf{y}})$ and, consequently, the column space of $\bar{\mathbf{X}}$ gets closer to the target $\mathbf{t}$ than that of $\mathbf{X}$; namely, the training accuracy increases in the next layer as stated in (16).

We prove Theorem 1 by leveraging the following lemma.

*Lemma:* Let $\mathbf{u} \in \mathrm{span}(\mathbf{X})$, $I \subseteq [N]$ be an index set neither empty nor universal, and $\mathbf{T}$ be a matrix, the columns of which form a subset of a basis of $\mathrm{span}(\mathbf{X})^\perp$. Then, $\rho(\mathbf{w}) \in \mathrm{span}(\mathbf{X})$ only if

$$\mathbf{w} \in \ker((\mathbf{Q}_I \mathbf{T})^T) \cap \ker((\mathbf{Q}_{I^c} \mathbf{T})^T) \tag{17}$$

for $\mathbf{w} \in \mathrm{span}(\mathbf{X})$ with $I(\mathbf{w} < 0) =: I$.

*Proof.* Let $\mathbf{w} \in \mathrm{span}(\mathbf{X})$, $\mathbf{T} = [\mathbf{v}_1, \ldots, \mathbf{v}_k]$, and $\rho(\mathbf{w}) \notin \mathrm{span}(\mathbf{X})$. Let either $\mathbf{w} \notin \ker((\mathbf{Q}_I \mathbf{T})^T)$ or $\mathbf{w} \notin \ker((\mathbf{Q}_{I^c} \mathbf{T})^T)$. If $\mathbf{w} \notin \ker((\mathbf{Q}_I \mathbf{T})^T)$ holds, then $(\mathbf{Q}_I \mathbf{v}_i)^T \mathbf{w} \neq 0$ for some $\mathbf{v}_i$. Therefore, $(\mathbf{Q}_{I^c} \mathbf{v}_i)^T \mathbf{w} \neq 0$ as

$$\mathbf{v}_i^T \mathbf{w} = (\mathbf{Q}_I \mathbf{v}_i)^T \mathbf{w} + (\mathbf{Q}_{I^c} \mathbf{v}_i)^T \mathbf{w} = 0. \tag{18}$$

However, $0 \neq (\mathbf{Q}_{I^c} \mathbf{v}_i)^T \mathbf{w} = \mathbf{v}_i^T \mathbf{Q}_{I^c} \mathbf{w} = \mathbf{v}_i^T \rho(\mathbf{w})$, which is a contradiction. On the other hand, if $\mathbf{w} \notin \ker((\mathbf{Q}_{I^c} \mathbf{T})^T)$ holds, then a contradiction exists in the same way.

*Proof of Theorem 1:* Equation (14) is proved by substituting $\hat{\mathbf{y}}$ and $[\hat{\mathbf{y}} - \mathbf{y}]$ (regarded as a matrix of size $N \times 1$) into $\mathbf{u}$ and $\mathbf{T}$, respectively, in lemma. Equation (15) is trivial as $\bar{\mathbf{X}}$ additionally contains a vector independent to all the column vectors of $\mathbf{X}$. To prove (16), since it suffices for $\rho(\hat{\mathbf{y}})$ to be comprised of an orthogonal vector, assume without loss of generality that $\rho(\hat{\mathbf{y}}) \in \ker \mathbf{X}^T$. Let $\mathbf{X} = \mathbf{U}\mathbf{\Sigma}\mathbf{V}^T$ be its SVD with $\mathbf{V} = [\mathbf{v}_1, \ldots, \mathbf{v}_d]$, $\mathbf{U} = [\mathbf{u}_1, \ldots, \mathbf{u}_d]$, $\mathbf{\Sigma} = \mathrm{diag}(\sigma_1, \ldots, \sigma_d)$. If $\bar{\mathbf{X}} = \bar{\mathbf{U}}\bar{\mathbf{\Sigma}}\bar{\mathbf{V}}^T$ is its SVD, then $\bar{\mathbf{V}} = [\bar{\mathbf{v}}_1, \ldots, \bar{\mathbf{v}}_d, \bar{\mathbf{v}}_{d+1}]$, $\bar{\mathbf{U}} = [\bar{\mathbf{u}}_1, \ldots, \bar{\mathbf{u}}_{d+1}]$, $\bar{\mathbf{\Sigma}} = \mathrm{diag}(\sigma_1, \ldots, \sigma_d, \rho(\hat{\mathbf{y}}))$, where $\bar{\mathbf{v}}_i = [\mathbf{v}_i^T, 0]^T$, $\bar{\mathbf{v}}_{d+1} = [\mathbf{0}^T, 1]^T$, $\bar{\mathbf{u}}_i = \mathbf{u}_i$, $\bar{\mathbf{u}}_{d+1} = \rho(\hat{\mathbf{y}})/\|\rho(\hat{\mathbf{y}})\|$ for $i = 1, \ldots, d$. By denoting $\hat{\mathbf{y}} = \mathbf{P}_\lambda \mathbf{y}$ and $\hat{\mathbf{y}}_{\text{next}} = \bar{\mathbf{P}}_\lambda \mathbf{y}$, where $\lambda$ is the ridge regularizer, we obtain

$$\mathbf{P}_\lambda = \sum_{i=1}^{d} \frac{\sigma_i^2}{\sigma_i^2 + \lambda} \mathbf{u}_i \mathbf{u}_i^T \tag{19}$$

$$\bar{\mathbf{P}}_\lambda = \mathbf{P}_\lambda + \mathbf{H}_\lambda \tag{20}$$

where $\mathbf{H}_\lambda := s_\lambda \rho(\hat{\mathbf{y}}) \rho(\hat{\mathbf{y}})^T$ with $s_\lambda := 1/(\rho(\hat{\mathbf{y}})^2 + \lambda)$. Also note that

$$\begin{aligned}
&\|\mathbf{y} - \mathbf{P}_\lambda \mathbf{y}\|^2 - \|\mathbf{y} - \bar{\mathbf{P}}_\lambda \mathbf{y}\|^2 \\
&= 2 s_\lambda \left(\rho(\hat{\mathbf{y}})^T \mathbf{y}\right)^2 - s_\lambda^2 \rho(\hat{\mathbf{y}})^2 \left(\rho(\hat{\mathbf{y}})^T \mathbf{y}\right)^2
\end{aligned} \tag{21}$$

due to the reason that $\mathbf{P}_\lambda^T \mathbf{H}_\lambda = 0$ by $\rho(\hat{\mathbf{y}}) \in \ker(\mathbf{X}^T)$. Now it is obvious that (21) is greater than 0, as $2 \geq s_\lambda \|\rho(\hat{\mathbf{y}})\|^2 = \|\rho(\hat{\mathbf{y}})\|^2 / (\|\rho(\hat{\mathbf{y}})\|^2 + \lambda)$. This completes the proof. ∎

The extension to the multiclass classification is trivial, and the proof above holds for arbitrary $\ell$ and $(\ell+1)$th layers.

### B. Layer-Wise Intra/Interclass Distance Dynamics in DAN

Descriptive features are, at least partially, implied by high interclass and low intraclass variances on them. We show that the deeper the layer of DAN is, the bigger the gap is between the variance of the interclass features and that of the intraclass ones. To this end, the expected intraclass and interclass distances, $w^{(\ell)}$ and $b^{(\ell)}$, respectively, at the $\ell$th layer of DAN are defined as follows:

$$\begin{aligned}
w^{(\ell)} &:= \left[ \|\mathbf{h}^{(\ell)} - \mathbf{h}'^{(\ell)}\|^2 \mid \mathbf{x}, \mathbf{x}' \in C_c \right] \\
b^{(\ell)} &:= \mathbb{E}\left[ \|\mathbf{h}^{(\ell)} - \mathbf{h}'^{(\ell)}\|^2 \mid \mathbf{x} \in C_c, \mathbf{x}' \in C_{c'} \right].
\end{aligned} \tag{22}$$

Here, $\mathbf{x}$ and $\mathbf{x}'$ are the independent identically distributed random input samples, $\mathbf{h}^{(\ell)}$ (resp. $\mathbf{h}'^{(\ell)}$) is the $\ell$th layer of DAN completed from $\mathbf{x}$ (resp. $\mathbf{x}'$), and $\mathbf{x} \in C_c$ (and $\mathbf{x}' \in C_{c'}$ similarly) indicates that $\mathbf{x}$ belongs to the $c$th class with $c \neq c'$ assumed.

As the regression error is believed to be normally distributed, we deliberate the following technical assumptions to simplify our formulation.

*Assumptions:* Let $\boldsymbol{\epsilon}^{(\ell)}$ (resp. $\boldsymbol{\epsilon}'^{(\ell)}$) represents the random vector modeling the prediction error

$$\boldsymbol{\epsilon}^{(\ell)} = \hat{\mathbf{y}}^{(\ell)} - \mathbf{y} \tag{23}$$

where $\hat{\mathbf{y}}^{(\ell)}$ is the regression output of $h^{(\ell)}$, $\mathbf{y} \in \{\mathbf{t}_1, \mathbf{t}_2, \ldots, \mathbf{t}_{N_c}\}$ is the true target of $\mathbf{x}$, and $\mathbf{t}_j = [t_{j,1}, \ldots, t_{j,j}, \ldots, t_{j,N_c}] \in \mathbb{R}^{N_c}$. Assume for arbitrary $j, c, c' \in \{1, \ldots, N_c\}$ with $c \neq c'$:

1) $\boldsymbol{\epsilon}^{(\ell)}$ and $\boldsymbol{\epsilon}'^{(\ell)}$ are of independent, given either $\mathbf{x}, \mathbf{x}' \in C_c$, or $\mathbf{x} \in C_c, \mathbf{x}' \in C_{c'}$;
2) the class-conditional distribution of prediction error remains the same, whichever class the input $\mathbf{x}$ belongs to; that is, $p_{\boldsymbol{\epsilon}^{(\ell)} | \mathbf{x} \in C_c} = p_{\boldsymbol{\epsilon}^{(\ell)} | \mathbf{x} \in C_{c'}}$;
3) $p_{\boldsymbol{\epsilon}^{(\ell)} | \mathbf{x} \in C_c} = \mathcal{N}(\mathbf{0}, \mathrm{diag}(\boldsymbol{\sigma}^{(\ell)2}(c)))$, that is, it follows multivariate normal distribution with variance $\mathrm{diag}(\boldsymbol{\sigma}^{(\ell)2}(c))$.

The same is assumed to hold for the test data with $\boldsymbol{\sigma}^{(\ell)}(c)$ replaced by $\boldsymbol{\sigma}_{\text{te}}(c)^{(\ell)}$. By Assumption 2), let us denote $p_{\boldsymbol{\epsilon}_j^{(\ell)} | x \in C_c}$ and $\boldsymbol{\sigma}^{(\ell)}(c)$ simply by $p^{(\ell)}$ and $\boldsymbol{\sigma}^{(\ell)}$, respectively.

Under these assumptions, we show that if the prediction error is sufficiently small, then the increment $\delta_w^{(\ell)} := w^{(\ell+1)} - w^{(\ell)}$ of the expected intraclass distance is relatively negligible compared to that of the expected interclass distance $\delta_b^{(\ell)} := b^{(\ell+1)} - b^{(\ell)}$. In detail, $\delta_b^{(\ell)}$ is greater than $t_{c,c}^2 + t_{c',c'}^2$ while $\delta_w^{(\ell)}$ is bounded by a factor of the variation of the prediction error $2 \| \sigma^{(\ell)} \|^2$. This implies that the gap $b^{(\ell)} - w^{(\ell)}$ becomes larger as the layer $\ell$ deepens, concluding the deeper layer of DAN gets more discriminative.

*Proposition 1:* We have

$$w^{(\ell+1)} = w^{(\ell)} + \delta_w^{(\ell)} \quad (24)$$
$$b^{(\ell+1)} = b^{(\ell)} + \delta_b^{(\ell)} \quad (25)$$

where

$$0 \leq \delta_w^{(\ell)} \leq 2 \| \sigma^{(\ell)} \|^2 \quad (26)$$

$$\delta_b^{(\ell)} \geq t_{c,c}^2 P_{c,c}^{(\ell)} + t_{c',c'}^2 P_{c',c'}^{(\ell)} + O\left(e^{\frac{-t_{\min}^2}{2\max\left(\sigma_c^{(\ell)}, \sigma_{c'}^{(\ell)}\right)^2}}\right) \quad (27)$$

as $\sigma^{(\ell)} \to 0$. Here, $P_{c,j}^{(\ell)} = \mathbb{P}(\epsilon_j^{(\ell)} > -t_{c,j} | \mathbf{x} \in C_c)$ denotes the probability that the prediction error at the $\ell$th layer is greater than $-t_{c,j}$, and $t_{\min} = \min_{c,j} |t_{c,j}|$. Moreover,

$$\delta_w^{(\ell)} = O\left(\sigma_c^{(\ell)2} + \sum_{j \neq c} \exp\left(-\frac{t_{\min}^2}{2\sigma_j^{(\ell)2}}\right)\right) \quad (28)$$

$$\delta_b^{(\ell)} \to t_{c,c}^2 + t_{c',c'}^2 \quad (29)$$

as $\sigma^{(\ell)} \to 0$.

*Corollary:* If the prediction error $\|\sigma^{(\ell)}\|$ is sufficiently small, the gap $b^{(\ell)} - w^{(\ell)}$ increases as the layer $\ell$ deepens with negligible variation on $w^{(\ell)}$.

*Proof of Proposition 1:* Here, we fix $\ell$ for clarity. Note

$$\|\mathbf{h}^{(\ell+1)} - \mathbf{h}'(\ell+1)\|^2 - \|\mathbf{h}^{(\ell)} - \mathbf{h}'(\ell)\|^2$$
$$= \sum_j \rho(\hat{y}_j)^2 - 2\rho(\hat{y}_j)^2 - 2\rho(\hat{y}_j)\rho(\hat{y}'_j) + \rho(\hat{y}'_j)^2 \quad (30)$$

$$\mathbb{E}[\rho(\hat{y}_j) | \mathbf{x} \in C_c] = t_{c,j} P_{c,j} + \sigma_j^2 p(t_{c,j}) \quad (31)$$
$$\mathbb{E}[\rho(\hat{y}_j)^2 | \mathbf{x} \in C_c] = (t_{c,j}^2 + \sigma_j^2) P_{c,j} + \sigma_j^2 t_{c,j} p(t_{c,j}) \quad (32)$$

which are obtained by change of variable on standard Gaussian integral $\int_{-t_{c,j}}^\infty \epsilon_j p(\epsilon_j) d\epsilon_j$ and $\int_{-t_{c,j}}^\infty \epsilon_j^2 p(\epsilon_j) d\epsilon_j$. Also, by Assumption 1)

$$\mathbb{E}[\rho(\hat{y}_j)\rho(\hat{y}'_j) | \mathbf{x} \in C_c, \mathbf{x}' \in C_{c'}]$$
$$= \mathbb{E}[\rho(\hat{y}_j) | \mathbf{x} \in C_c] \mathbb{E}[\rho(\hat{y}'_j) | \mathbf{x}' \in C_{c'}]. \quad (33)$$

Using these, we obtain

$$\delta_w^{(\ell)} = 2\sum_j t_{c,j} P_{c,j}^{(\ell)} \left(1 - P_{c,j}^{(\ell)}\right)$$
$$+ \sigma_j^{(\ell)2} \Big[ P_{c,j}^{(\ell)} \left(1 - 2t_{c,j} p^{(\ell)}(t_{c,j})\right)$$
$$+ t_{c,j} p^{(\ell)}(t_{c,j}) - \sigma_j^{(\ell)2} p^{(\ell)}(t_{c,j})^2 \Big] \quad (34)$$

and

$$\delta_b^{(\ell)} = \sum_j \Big[ t_{c,j}^2 P_{c,j}^{(\ell)} - 2t_{c,j} t_{c',j} P_{c,j}^{(\ell)} P_{c',j}^{(\ell)} + t_{c',j}^2 P_{c',j}^{(\ell)} \Big]$$
$$+ \sigma^{(\ell)2} \Big[ t_{c,j} p^{(\ell)}(t_{c,j}) + t_{c',j} p^{(\ell)}(t_{c',j})$$
$$+ P_{c,j}^{(\ell)} \left(1 - 2p^{(\ell)}(t_{c',j}) t_{c,j}\right)$$
$$+ P_{c',j}^{(\ell)} \left(1 - 2p^{(\ell)}(t_{c,j}) t_{c',j}\right)$$
$$- 2\sigma_j^{(\ell)2} p^{(\ell)}(t_{c,j}) p^{(\ell)} t_{c',j} \Big]. \quad (35)$$

By upper bounds for Gaussian tail

$$P_{c,j} < \frac{\sqrt{2}}{\sqrt{\pi}\left(\sqrt{8/\pi + (t_{c,j}/\sigma_j)^2} + |t_{c,j}|/\sigma_j\right)} \exp\left(-\frac{t_{c,j}^2}{2\sigma_j^2}\right) \quad (36)$$

for $j \neq c$, which implicates that $P_{c,j}$, $1 - P_{c,j}$, and $p(t_{c,j})$ for $j \neq c$, $j = c$, and all $j$, respectively, are all $O(\exp(-1/\sigma_j^2))$ as $\sigma_j \to 0$. Applying these on (34) and (35), it proves (27)–(29). Meanwhile, (26) is proved by observing $|\rho(\hat{y}_j) - \rho(\hat{y}'_{j'})| \leq |\hat{y}_j - \hat{y}'_j|$ and the expectation of $|\hat{\mathbf{y}} - \hat{\mathbf{y}}'|$, provided that $\mathbf{x}, \mathbf{x}' \in C_c$, is $2\sum_j \sigma_j^2$. This completes the proof. ∎

Noting the fact that assumptions hold on test data with $\sigma^{(\ell)}$ replaced by $\sigma_{te}^{(\ell)}$, the layer-wise inter/intraclass distance dynamics described in Proposition 1 applies at test phase as well. This is experimentally verified as shown in Figs. 5 and 6.

## V. EXPERIMENTS AND DISCUSSION

We recapitulate, in this section, the DAN/K-DAN generalization performance for pattern classification. We analyze and compare DAN/K-DAN to the present FC S-DNNs, and the BP-trained networks, for example, S-DNN$_{BP}$, MLP, and other DNNs.

### A. Benchmarking Datasets

Our experiments recruit datasets of varying domains, including faces, digits, natural objects, and structured features.

1) FERET [36] consists of a training set FA with a single face image for each of 1196 subjects; and 4 probe sets with pose, expression, illumination, and time-span variations, namely, FB, FC, DUP I, and DUP II— each containing a summation of 1195, 194, 722, and 234 images of size $128 \times 128$ pixels.
2) MNIST [37] is composed of 70 000 handwritten digits, digit 0 to digit 9, each is of size $28 \times 28$ pixels. Compliant to the predetermined evaluation protocol, the first 60 000 images are apportioned for training, and the rest for testing.
3) CIFAR10 [38] reposits 60 000 color images of each $32 \times 32$ pixels for 10 natural objects. The training and the testing set capacities are of 50 000 and 10 000, respectively.
4) Tiny ImageNet [39] refers to the ImageNet subset [40] furnished with a reduced labeled image set of 200 classes. Each class is sampled with 500 training, 50 validation, and 50 testing images

TABLE III
SPECIFICATION SUMMARY FOR UCI DATASETS

| UCI DATASETS | # TRAIN / # TEST | # FEA. | # CLS. | # TRAIN / CLS. (ON AVERAGE) |
|---|---|---|---|---|
| AUS. CREDIT | 460 / 230 | 7 | 2 | 230 |
| MUSHROOM | 1,500 / 6,624 | 112 | 2 | 750 |
| CONNECT-4 | 50,000 / 17,557 | 126 | 3 | 16,666 |
| GLASS | 142 / 72 | 9 | 6 | 23 |
| SATIMAGE | 4,435 / 2,000 | 36 | 6 | 739 |
| SHUTTLE | 43,500 / 14,500 | 9 | 7 | 6,214 |
| USPS | 7,291 / 2,007 | 256 | 10 | 729 |
| LETTER | 13,333 / 6,667 | 16 | 26 | 512 |

TABLE IV
PARAMETER CONFIGURATION FOR 2-FFC$_{\text{PCA}}$ FEATURES

| DATASET | PCA FILTER SET | # 1- / 2-FFC FILTERS | GRID BLK |
|---|---|---|---|
| FERET | $7 \times 7$ | - / 64 | $8 \times 8$ |
| MNIST | $7 \times 7$ | 8 / 64 | $[4, 2, 1]^+$ |
| CIFAR10 | $5 \times 5, 7 \times 7$ | 16 / 256 | $[4, 2, 1]^+$ |

Note: $^+$ refers to overlapping spatial pyramid pooling (SPP) [46] feature encoding.

composing the training/validation/testing sets of 100 000/10 000/10 000 images, respectively. Note that the original ImageNet contains 1.2-million of images with 1000 classes queried from the Internet, and the images are with variable appearances, positions, viewpoints, poses, background clutters, and occlusions. In addition, the image size is intentionally shrunk to only $64 \times 64$ pixels from $256 \times 256$ for visual recognition challenge. As the ground truths for the testing images are not provided, our evaluation is reported based on the validation set.

5) Apart from imagery datasets our experiments employ also the UCI machine learning repositories hand-engineered with structured features [41]. The data specifications for the preselected two-class and multiclass problems are summarized in Table III.

### B. Implementation Summary

For each of the imagery datasets, except Tiny ImageNet, a PCA filter ensemble is trained to render the onefold and the twofold PCA filter-to-filter convolution descriptors, that is, 1-FFC$_{\text{PCA}}$ and 2-FFC$_{\text{PCA}}$ [43]. Following that we learn DAN/K-DAN with $\mathcal{L}$-layer from these features. Table IV reveals the parameter setting for the 1-FFC$_{\text{PCA}}$ and 2-FFC$_{\text{PCA}}$ feature extraction and encoding stage, including PCA filter size (before 2-FFC), the number of 1- and 2-FFC PCA filters, and the grid-partition for block-wise histogram feature encoding, or optionally the overlapped spatial pyramid pooling (SPP) [46]. In addition to that we replicate the PCANet features according to the parameters recommended in [10] for performance comparison. For small-scale UCI machine learning repositories (of which the explicit feature encoding is inapplicable), we streamline the K-DAN structure to that with the last FT layer removed, known as K-DAN$_{\text{Trim}}$. This reduces the number of hyper-parameters from the initial four, that is, $\lambda^{(\ell)}$, $\gamma^{(\ell)}$, $\lambda^{(\ell)}_{\text{FT}}$, and $\beta_{\text{FT}}$, to only two, that is, $\lambda^{(\ell)}$ and $\gamma^{(\ell)}$.

Our experiments recruit no additional manipulation to preprocess the images. We report the DAN/KDAN performance in terms of rank-1 classification accuracy (%). Since the DAN/K-DAN construction is built layer by layer, we explore the layer-wise performance and the best-performing layer are remarked for comparison. However, for the end-to-end trained networks, e.g., MLP, the layer-wise performance is inaccessible.

### C. Parameter Configuration

In place of grid-searching for the best parameter setting for each individual layer $\ell$, the layer-wise hyper-parameters of DAN/K-DAN are coarsely fine-tuned across $\mathcal{L}$ layers with respect to the validation sets (to be remarked in the following sections). To be specific, we simplify the trivial tuning task by fixing the layer-wise parameters to be similar for all layers, from the first to the last. For result replicability, we list the parameter configuration for each benchmarking dataset in Table V.

### D. Performance Analysis

We degeneralize DAN/K-DAN into its basic configurations for a pilot study, followed by a relearnability test. Along with that the DAN/K-DAN theories delivered in Section IV are validated in this section. As DAN/K-DAN are analogous algorithmically, our analysis considers only DAN.

*1) Basic Configuration and Relearnability Analysis:* The DAN construction (as portrayed in Fig. 3) is degeneralized into that with/without ReLU and the FT layer. We carefully fine-tune DAN in the first layer to scrutinize the extent to which the deep construction improves the layer-wise performance. To investigate if the relearned features lie in the feature domain, the built-in regression classifier is replaced by the NN classifier with the Euclidean distance metric. We index all configurations from I to V as follows:

I. linear DAN with no ReLU and no FT layer;
II. linear DAN with an FT layer;
III. complete DAN with ReLU and FT;
IV. carefully fine-tuned DAN;
V. DAN with NN classifier, replacing the built-in regression classifier.

For the ten-layer DAN learned from the 2-FFC$_{\text{PCA}}$ features pre-extracted for the FERET FA images, Table VI discloses that the linear DAN (configuration I) only learns in the first two layers, capping at 94.44% evaluated on the DUP II probe set. However, appending a power-regularized RR-based FT layer to the linear DAN, or configuration II, improves the layer-wise performance from 44.44% to 95.73%, navigating from the first to the deepest layer. The primary reason leads to the drastic performance drop is that the FT layer trains the built-in regression classifier from the power-regularized (distorted) prediction outputs without the 2-FFC$_{\text{PCA}}$ features. For a sufficiently deep network, we discern that the stack of power-regularized prediction outputs yields an expressive feature set for the FT learning stage. On top of that, we discern that ReLU and FT are complementary to each other. The nonlinear DAN with ReLU and the FT layer escalates the baseline performance of the 2-FFC$_{PCA}$ features from 83.73% to 97.01% in the third layer (refer to configuration III).

TABLE V
PARAMETER CONFIGURATION

| DATASET | DAN | | | K-DAN | | | |
|---|---|---|---|---|---|---|---|
| | $\lambda^{(\ell)}$ | $\lambda^{(\ell)}_{FT}$ | $\beta_{FT}$ | $\gamma^{(\ell)}$ | $\lambda^{(\ell)}$ | $\lambda^{(\ell)}_{FT}$ | $\beta_{FT}$ |
| FERET | 0.1 | 0.1 | 0.5 | 0.01 | 0.001 | 0.001 | 0.5 |
| MNIST | 1.0 | 0.00001 | 0.5 | 0.1 | 0.01 | 0.01 | 0.5 |
| CIFAR10 | 6.0 | 0.1 | 1.0 | 0.001 | 0.01 | 0.00001 | 0.6 |
| TINY IMAGENET | 10 | 0.1 | 1.0 | | | | |

| UCI DATASETS | K-DAN$_{TRIM}$ | | UCI DATASETS | K-DAN$_{TRIM}$ | |
|---|---|---|---|---|---|
| | $\lambda^{(\ell)}$ | $\gamma^{(\ell)}$ | | $\lambda^{(\ell)}$ | $\gamma^{(\ell)}$ |
| AUS. CREDIT | 0.0001 | 0.009 | SATIMAGE | 0.01 | 0.2 |
| MUSHROOM | 0.4 | 0.9 | SHUTTLE | 0.01 | 0.9 |
| CONNECT-4 | 0.01 | 0.2 | USPS | 0.001 | 0.01 |
| CLASS | 0.1 | 0.3 | LETTER | 0.1 | 0.25 |

TABLE VI
PERFORMANCE ANALYSIS ON FERET DUP II PROBE SET, IN TERMS OF CLASSIFICATION ACCURACY (%)

| DESCR. | FERET DUP II | | | | |
|---|---|---|---|---|---|
| 2-FFC$_{PCA}$ [43] (TCSVT, 2019) | 83.76 | | | | |
| DAN$_\ell$ / NET. CONF. | I | II | III | IV | V |
| DAN$_1$ + 2-FFC$_{PCA}$ | 94.02 | 44.44 | 95.73 | 96.15 | 95.30 |
| DAN$_2$ + 2-FFC$_{PCA}$ | **94.44** | 62.82 | 96.15 | **97.01** | 96.15 |
| DAN$_3$ + 2-FFC$_{PCA}$ | 94.44 | 79.06 | **97.01** | **97.01** | **97.01** |
| DAN$_4$ + 2-FFC$_{PCA}$ | 94.44 | 85.90 | 96.58 | 96.15 | 96.58 |
| DAN$_5$ + 2-FFC$_{PCA}$ | 94.44 | 90.60 | 96.15 | 95.30 | 95.73 |
| DAN$_{10}$ + 2-FFC$_{PCA}$ | 94.44 | **95.30** | 95.30 | 95.30 | 95.73 |

In the meantime, we observe that the carefully fine-tuned DAN, that is, configuration IV, accomplishes only 96.15% in the first layer. If it is deepened further while freezing the hyperparameter settings unchanged, its classification performance is progressed to 97.01%. This discloses that DAN does refine its representation by means of relearning at the time growing a layer deeper. For the DAN equipped with the NN classifier, or configuration IV, the ReLU-ed predictive outputs are revealed interpretable as the relearned features since the performance by pair-matching is unaffected, comparing to that of configuration III. This underscores that the relearned features are applicable also to other succeeding manipulations, e.g., metric learning. However, this is beyond the coverage of this paper. On the other hand, since the 2-FFC$_{PCA}$ feature dimension $d = 131\,072$ (refer to the parameter configurations in Table IV), and the FT output dimension $N_c = 1096$, such that $N_c < d$, DAN performs also feature compression.

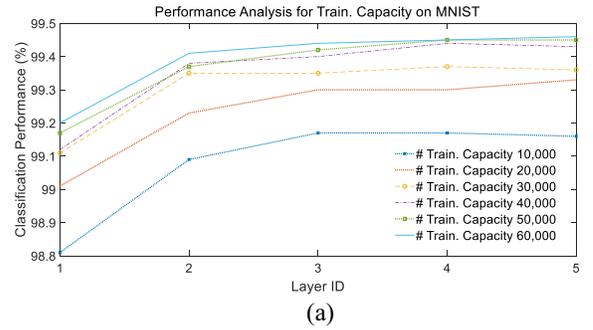

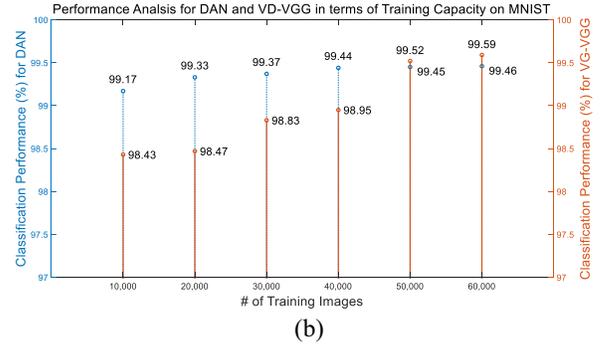

Fig. 4. Performance analysis on training capacity for (a) DAN, and the (b) BP-trained VD-VGG using MNIST.

*2) Training Capacity Analysis:* We switch from FERET to MNIST owing to the larger training capability available for the empirical analyses. We sample six training subsets with 10 000 to 60 000 images to learn for each subset a five-layer DAN with respect to the pre-extracted 1- and 2-FFC$_{PCA}$ features. Similar to other learning-based models, we disclose in Fig. 4(a) that the DAN performance is proportionate to the number of training samples. Our empirical results show that the training subset with 10 000 samples is sufficient to learn an outperforming DAN. The DAN trained from this subset achieves an accuracy of 99.17%, prevailing over the BP-trained very deep VGG network of 16 layers (VD-VGG-16) [56] with only 98.43%. The same phenomenon is observed for capacities up to 40 000 examples. However, the VD-VGG-16 performance is advanced to 99.59% for the counterpart learned from the entire training set. This shows that DAN gains performance advantage over the BP networks, particularly when the training capacity is limited.

*3) Theory Analysis:* For theory analysis, a summation of 1000 random training and testing samples are drawn with replacement from the MNIST dataset to investigate the intra/interclass variations in terms of Euclidean distance. Our empirical finding in Fig. 5 reflects the Proposition 1 in Section IV that the interclass distance dynamically increases whereas the intraclass distance is mildly perturbed, each time navigating a layer deeper. We, therefore, substantiate that the deeply stacked DAN (and K-DAN) works by deviating the interclass samples, while preserving the intraclass distribution.

To validate the assertion made by Proposition 1, the physical intra/interclass distances, i.e., $w^{(\ell)}$ and $b^{(\ell)}$ measured as in (22) based on our empirical samples, are compared to the theoretical intra/interclass distances $w_{\text{theo}}^{(\ell)}$ and $b_{\text{theo}}^{(\ell)}$ as in (24) and (25), respectively. Note $w^{(1)} = w_{\text{theo}}^{(1)}$ and $b^{(1)} = b_{\text{theo}}^{(1)}$ for the first layer. Fig. 6 shows that our theoretical simulations approximate the physical ones.

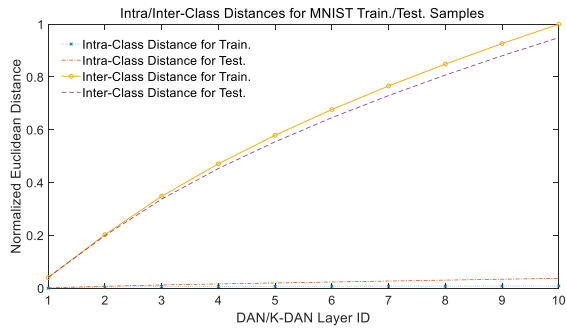

Fig. 5. Intra/interclass distances for MNIST training and testing samples.

### E. Performance Comparison and Discussion

We distinguish DAN/K-DAN from the primary S-DNN and S-DNN$_{\text{BP}}$ counterparts in this section. For a thorough analysis and comparison, the three-hidden-layer MLP with ReLU and batch normalization [54] (MLP$_{\text{ReLU-BN}}$) is trained for each dataset to relieve the overfitting problem. The number of hidden nodes for each layer is to be disclosed in the following sections accordingly. Other configurations are: cross-entropy as the MLP loss function; a constant weight decay and momentum of 0.0005 and 0.9; the number of epochs is set to 80–100 for FERET, MNIST, and CIFAR, but 50 for UCI datasets; and a learning decay of 0.1 for each 10 epochs.

*1) FERET:* Owing to the rigorous evaluation protocol (with merely a single image per subject in the FA training set), the S-DNNs reviewed in Section I disregard this dataset for face recognition analysis. We, therefore, only compare DAN/K-DAN to the three-hidden-layer MLP$_{\text{ReLU-BN}}$ (equipped with 500 nodes in each layer) and KDCN [25] in Table VII, where the best-performing layer for DAN, K-DAN, and KDCN is parenthesized. Since BP networks require a large training capacity of target specific images (refer to Fig. 4), we witness that the FA-learned MLP$_{\text{ReLU-BN}}$ produces a very high misclassification rate across all probe sets. Despite of being better than MLP$_{\text{ReLU-BN}}$, the modularly learned K-DAN (from raw image pixels) is attested to be underperformed. This is due to the wide variations in the testing distribution incurred by poses, facial expressions, illumination conditions, time-span, and other disturbances, opposing to the only frontal face in FA.

On the contrary, DAN/K-DAN gets rid of the variation issue by learning from the pre-extracted BSIF [42], PCANet [10], and 2-FFC$_{\text{PCA}}$ [43] features. We discern that with DAN, K-DAN, or KDCN, the baseline performance for PCANet and 2-FFC$_{\text{PCA}}$ in particular exhibits a vast improvement over the four probe sets. In a nutshell, the DAN/KDAN performance dominates KDCN for the less discriminative descriptor, that is, BSIF as in this case. We observe that KDCN halts from learning immediately after the first layer; while DAN/K-DAN continue relearning in each layer as it deepens. We strongly believe the DAN performance with the 2-FFC$_{\text{PCA}}$ features, that is, 98.69% on average, is the best at present. On the other hand, as the PCANet and the 2-FFC$_{\text{PCA}}$ features reside in high feature space (each consisting of 131 072 dimensions yielded based on Table IV), our results reveal that the indefinite projection by RBF for K-DAN and *K*-DCN offers no benefit. However, K-DAN is shown outperforming DAN for the BSIF features of 16 284 dimensions. Note that by adopting DUP II as the validation set, DAN/K-DAN are, respectively, trained with 5/6 layers for BSIF, 2/4 layers for PCANet, and 3/2 layers for 2-FFC$_{\text{PCA}}$.

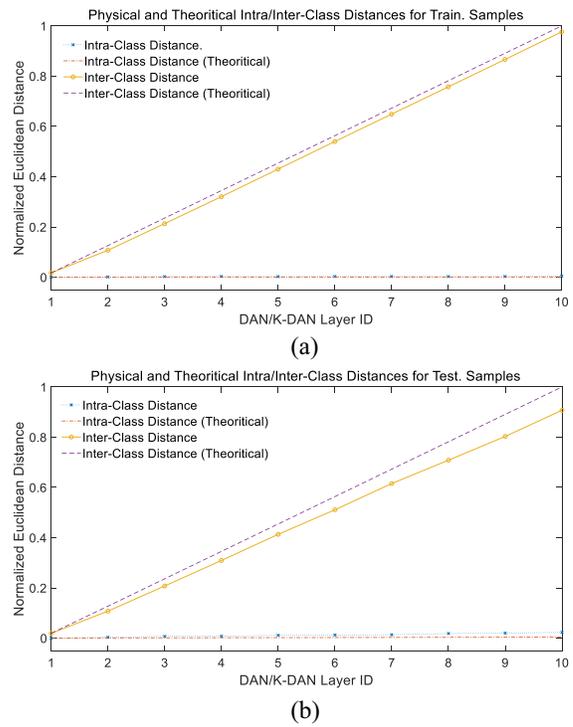

Fig. 6. Physical and theoretical intra/interclass distances for (a) training and (b) testing samples.

TABLE VII
PERFORMANCE COMPARISON FOR DAN/K-DAN, KDCN, AND MLP ON FERET, IN TERMS OF CLASSIFICATION ACCURACY (%)

| DESCR. | FB | FC | DUP I | DUP II (VALID.) | MEAN |
|---|---|---|---|---|---|
| MLP$_{\text{ReLU-BN}}$ (500-500-500) | 61.17 | 10.31 | 25.07 | 18.80 | 28.84 |
| K-DAN (Raw Image Pixels) | 83.68 (4) | 80.93 (4) | 65.80 (4) | 63.25 (4) | 73.42 |
| BSIF [42] (ICPR, 2012) | 93.47 | 69.07 | 71.75 | 59.40 | 73.42 |
| BSIF + KDCN [25] | 99.25 (1) | 98.97 (1) | 87.67 (1) | 85.47 (1) | 92.84 |
| BSIF + DAN | 99.58 (6) | 100 (2) | 92.66 (7) | 89.74 (7) | 95.50 |
| BSIF + K-DAN | 99.58 (6) | 100 (4) | 93.49 (6) | 91.03 (6) | 96.03 |
| PCANet [10] (TIP, 2015) | 95.56 | 99.48 | 86.29 | 84.62 | 91.49 |
| PCANet + KDCN [25] | 99.75 (2) | 100 (1) | 96.82 (2) | 95.30 (2) | 97.97 |
| PCANet + DAN | 99.75 (2) | 100 (1) | 97.92 (2) | 96.15 (2) | 98.46 |
| PCANet + K-DAN | 99.83 (3) | 100 (2) | 97.78 (4) | 96.15 (4) | 98.44 |
| 2-FFC$_{\text{PCA}}$ [43] (TCSVT, 2019) | 95.65 | 99.48 | 86.57 | 83.76 | 91.36 |
| 2-FFC$_{\text{PCA}}$ + KDCN [25] | 99.75 (1) | 100 (1) | 96.82 (2) | 95.30 (2) | 97.97 |
| 2-FFC$_{\text{PCA}}$ + DAN | 99.83 (2) | 100 (1) | 97.92 (3) | 97.01 (3) | **98.69** |
| 2-FFC$_{\text{PCA}}$ + K-DAN | 99.75 (1) | 100 (1) | 97.51 (2) | 97.01 (2) | 98.57 |

TABLE VIII
PERFORMANCE COMPARISON FOR DAN/K-DAN AND OTHER COUNTERPARTS ON MNIST, IN TERMS OF CLASSIFICATION ACCURACY (%)

| NET. DESCR. (NON-BP) | ACC. (%) | NET. DESCR. (BP-TRAINED) | ACC. (%) |
|---|---|---|---|
| T-DSN [27] (TPAMI, 2013) | 98.79 | MLP$_{\text{ReLU-BN}}$ (2000-2000-2000) | 97.96 |
| S-ELM [29] (TC, 2015) | 98.51 | DBN [16] (Neural Comput. 2006) | 98.75 |
| AE-S-ELM [29] (TC, 2015) | 98.89 | CDBN [17] (ICML, 2009) | 99.18 |
| H-ELM [30] (TNNLS, 2016) | 99.13 | DBM [19] (JMLR, 2010) | 98.83 |
| DF [32] (IJCAI, 2017) | 98.96 | SAE [19] (JMLR, 2010) | 98.60 |
| RBM-GI [35] (TSMC, 2017) | 97.58 | SDAE [19] (JMLR, 2010) | 98.72 |
| 1-2-FFC$_{\text{PCA}}$ [43] (TCSVT, 2019) | 99.03 | VD-VGG w/o D. Aug. [56] (ICLR, 2015) | 99.59 |
| 1-2-FFC$_{\text{PCA}}$ + DAN | 99.46 [5] | R-NN w/o D. Aug. [45] (ICML, 2013) | 99.43 |
| 1-2-FFC$_{\text{PCA}}$+ K-DAN | 99.51 [3] | R-NN w/ D. Aug. [45] (ICML, 2013) | **99.79** |

TABLE IX
PERFORMANCE COMPARISON FOR DAN/K-DAN, AND OTHER COUNTERPARTS ON CIFAR10, IN TERMS OF CLASSIFICATION ACCURACY (%)

| NET. DESCR. (NON-BP) | ACC. (%) | NET. DESCR. (BP-TRAINED) | ACC. (%) |
|---|---|---|---|
| D-ELM [47] (Neurocomputing, 2016) | 56.00 | MLP$_{\text{ReLU-BN}}$ (2000-2000-2000) | 58.95 |
| DF [32] (IJCAI, 2017) | 63.37 | DBN [16] (Neural Comput. 2006) | 62.20 |
| **A** : PCANet [10] (TIP, 2015) | 65.10 | **C** : VD-VGG w/o D. Aug. [56] (ICLR, 2015) | 87.57 |
| **B** : 1-2-FFC$_{\text{PCA}}$ [43] (TCSVT, 2019) | 63.38 | **C** + 1-NN | 87.01 |
| **A** + DAN$_{\text{SVM}}$ | 76.66 [5] | **C** + DAN | 87.78 [4] |
| **B** + DAN$_{\text{SVM}}$ | 79.03 [4] | **C** + K-DAN | 87.97 [3] |
| **A** + **B** + DAN$_{\text{SVM}}$ | 80.15 [3] | FMP-Net [47] (arXiv, 2015) | **96.53** |
| | | Large ALL-CNN [48] (ICLR, 2015) | 95.59 |
| | | LSUV-Net [49] (ICLR, 2016) | 94.16 |

*2) MNIST:* For performance analysis, we learn DAN/K-DAN with five layers from the training/validation sets of 50 000/10 000 examples. In addition to the S-DNN counterparts, we compare DAN/K-DAN to S-DNN$_{\text{BP}}$, MLP$_{\text{ReLU-BN}}$, and the two remarkable CNNs, that is, very deep VGG of 16 layers (VD-VGG) [56] and the regularized neural network (R-NN) [45]. Table VIII shows that the DAN/K-DAN learned based on the 1-2-FFC$_{\text{PCA}}$ features, that is, the composition of 1-FFC$_{\text{PCA}}$ and 2-FFC$_{\text{PCA}}$ features, attain 99.46% and 99.51% of accuracies, extended from the baseline of 99.03%. This attest that both DAN/K-DAN are the best-performing ones compared with other S-DNN counterparts. It is reported that S-ELM and AE-S-ELM [29] are learned with 650 and 700 layers in depth; DF [32] learns for every layer an ensemble of random decision trees; in place of one-shot solution like DAN/K-DAN, T-DSN [27], and RBM-GI [35] are trained iteratively, despite of BP FT is not exercised; and the remaining networks are S-DNN$_{\text{BP}}$ instances, namely DBN [16], CDBN [17], DBM [19], SAE [19], and SDAE [19].

To date, the least generalization error for MNIST is archived to be 0.21%, equivalent to a classification accuracy of 99.79%, by R-NN [45]. Rather than a single network, R-NN learns a bag of five with aggressive data augmentation and its final accuracy is determined via voting. However, we discern that its accuracy without data augmentation shrinks to 99.43%, outperformed by that of DAN/K-DAN marginally. Our analysis in the preceding section reveals also that the single VD-VGG network achieves an impressive accuracy of 99.59%, without data augmentation applied.

*3) CIFAR10:* Our experiments learn for CIFAR10 the 6-layer DAN/K-DAN from the random training/validation sets with 40 000/10 000 examples. To the best of our knowledge, the only two non-BP S-DNNs evaluating on CIFAR10 are D-ELM [47] and DF [32]. With a minimal performance of 63.37%, Table IX summarizes that DF stands out from non-BP D-ELM and the BP networks, including the 3-layer MLP$_{\text{ReLU-BN}}$ (designated with 2000 nodes for each hidden layer) and DBN [16].

Different from the datasets discussed earlier, we opt DAN for the linear SVM classifier, in place of the RR building block. By fixing the SVM penalty parameter to 0.1, we learn the DAN$_{\text{SVM}}$ from the PCA-compressed 1-2-FFC$_{\text{PCA}}$ composite features with only 4000 dimensions. We proves that DAN$_{\text{SVM}}$ improves the baseline accuracy by 16%, progressing from 63.38% to 79.03%, even with the PCA-compressed features. In addition to that we demonstrate that DAN$_{\text{SVM}}$ achieves 76.66%, extended from an accuracy of 65.10% for the PCA-compressed PCANet features of 4000 dimensions derived from $3 \times 3$ and $5 \times 5$ filters based on [10]. On top of that the composition of the PCA-compressed 1-2-FFC$_{\text{PCA}}$ and PCANet features offers an accuracy of 80.15% in the third layer (refer to the row of **A** + **B** in Table IX).

To bridge DAN and K-DAN to the BP-trained networks, we pretrain the 16-layer VD-VGG [56] with the softmax classifier. We extract the features learned at the last FC layer for our subsequent analyses using the 1-NN classifier, DAN, and K-DAN. We attest that the DAN/K-DAN-trained based on the prelearned VD-VGG features outperforms 1-NN and the commonly used softmax classifier. This suggests DAN/K-DAN to be practiced in transfer learning [58], such that DAN or K-DAN is analytically trained as an auxiliary classifier in place of softmax. We scrutinize this using the Tiny ImageNet dataset in the following section.

The three top-ranked BP-trained DNNs for CIFAR10 are the fractional max-pooling network (FMP-Net) [47], the large for all CNN (large ALL-CNN) [48], and the layer-sequential unit-variance network (LSUV-Net) [49], each of which achieves 96.53%, 95.59%, and 94.16%, respectively. The two common grounds for these top-performing DNNs are as follows.

TABLE X
Performance Comparison for DAN/K-DAN and Other Classifiers on Tiny ImageNet, in Terms of Top-1 Accuracy (%)

| Net. Descr. | Acc. (%) |
|---|---|
| VD-VGG-19 [56] + SoftMax (training from random initialization) | 59.79 |
| Pre-trained VD-VGG-19 [56] + 1-NN | 55.44 |
| Pre-trained VD-VGG-19 [56] + FC-SoftMax | 67.59 |
| Pre-trained VD-VGG-19 [56] + DAN | 69.34 [8] |
| Pre-trained ResNet-156 [57] + 1-NN | 63.29 |
| Pre-trained ResNet-156 [57] + FC-SoftMax | 76.17 |
| Pre-trained ResNet-156 [57] + DAN | 76.84 [8] |
| Pre-trained VD-VGG-19 [56] + Pre-trained ResNet-156 [57] + DAN | **78.02** [5] |

TABLE XI
Performance Comparison for K-DAN$_{\text{Trim}}$ and Other Counterparts, in Terms of Classification Accuracy (%), for UCI Datasets

| UCI Datasets | Net. Descr. | | | | |
|---|---|---|---|---|---|
| | H-ELM [30] | DBN [16] | RBM-GI [35] | MLP$_{\text{ReLU-BN}}$ | K-DAN$_{\text{Trim}}$ |
| Aus. Credit | 86.96 | - | - | 75.22 (80-80-80) | **87.22** [5] |
| Mushroom | **100** | **100** | **100** | 61.94 (80-80-80) | **100** [5] |
| Connect-4 | - | 83.85 | 76.06 | 86.27 (200-200-200) | **91.01** [4] |
| Class | 76.39 | - | - | 22.22 (50-50) | **77.50** [3] |
| Satimage | **90.90** | 85.30 | 84.75 | 89.40 (500-500-500) | 89.94 [4] |
| Shuttle | - | 96.03 | 91.55 | 99.24 (200-200-200) | **99.93** [6] |
| USPS | 96.76 | 94.67 | 91.38 | 97.91 (500-500-500) | **98.10** [4] |
| Letter | 95.82 | 62.05 | 74.45 | 96.68 (200-200-200) | **97.45** [4] |

1) Instead of only one, multiple networks are learned greedily on different initializations and configurations. For example, FMP-Net is learned with summation of 100 networks.
2) Data augmentation is of mandatory to gain training diversity for performance improvement.

Note that, both FMP-Net and LSUV-Net are renamed for self-explanatory convenience.

*4) Tiny ImageNet:* This section employs two very deep CNNs prelearned from the complete ImageNet dataset implemented by MatConvNet [52]—the 19-layer VD-VGG (VD-VGG-19) [56], and the 156-layer ResNet (ResNet-156) [57]. We affix these CNNs with three different classifiers, including the NN with Cosine distance metric (1-NN), the softmax classifier with a single FC layer, and the 8-layer DAN, of which both softmax classifier and DAN are trained from the VD-VGG-19 and RestNet-156.

Interestingly, Table X summarizes that DAN prevails over 1-NN and the softmax classifier for both CNN-extracted features, in terms of top-1 classification accuracies. The concatenation of the VD-VGG-19 and the ResNet-156 features further improves the accuracies from 69.34% (for VD-VGG-19) and 76.84% (for RestNet-156) to 78.08%. In the meantime, we discern that VD-VGG-19 accomplish a relatively poor accuracy of 59.79%, if it is retrained from random initializations on the Tiny ImageNet training set. This discloses that the non-BP DAN also relearns from the CNN-learned features, and it is therefore a good option to the softmax classifier in the transfer learning practice.

*5) UCI Hand-Engineered Datasets:* In place of K-DAN, we apply K-DAN$_{\text{Trim}}$ to untangle the UCI hand-engineered problems. Each UCI dataset is reshuffled for ten trials in our experiments, and the two hyper-parameters, i.e., $\lambda^{(\ell)}$ and $\gamma^{(\ell)}$, are empirically set as in Table V. We summarize the average classification rate over all trials in Table XI, along with performance summary for H-ELM [30], DBN [16], RBM-GI [35], and MLP$_{\text{ReLU-BN}}$. The MLP structure for each dataset, either of two or three hidden layers, is parenthesized.

Although with the FT layer withdrawn, we attest that K-DAN$_{\text{Trim}}$ outshines other counterparts on all datasets, except for Satimage. Moreover, our experimental results disclose that K-DAN outperforms MLP$_{\text{ReLU-BN}}$ on the whole. For small-scale datasets, that is, the Glass dataset with 23 training images for each class, MLP$_{\text{ReLU-BN}}$ is revealed inferior achieving only 22.22%; whereas K-DAN$_{\text{Trim}}$ is appraised to be 77.50%. For other larger datasets, especially Connect-4 with 16 666 images per class, K-DAN$_{\text{Trim}}$ also exhibits its superiority over both MLP$_{\text{ReLU-BN}}$ and DBN. This ascertains that a simplistic analytic network like K-DAN$_{\text{Trim}}$ is a viable alternative to BP-based MLP and DBN.

*F. Comparison to Other BP-Trained Networks*

We compare DAN/KDAN to the most influential BP-optimized CNNs from three perspectives: the most primitive performance (with only a single network trained for pattern classification and recruiting no data augmentation); the computational complexity (the number of trainable parameters); the CPU training time (in h); and also the CPU inference time (in s). Following MatConvNet [52], we retrain AlexNet [54], network-in-network (NIN) [55], and residual network of depth 20 (ResNet-20) [57] on CIFAR10, aside from the aforementioned MLP$_{\text{ReLU-BN}}$ and VD-VGG of 16 layers [56]. Our implementation runs on an Intel Core i7-6850K @3.60GHz CPU and a NVidia GeForce GTX 750 GPU.

For a fair comparison, we first learn all BP networks by GPU for parameter tuning, and each is further retrained with the prelearned parameters by CPU for training and inference time. We observe from Fig. 7 that:

1) although MLP$_{\text{ReLU-BN}}$ learns a massive parameter set by BP (approximately 14.14M), its performance is limited to only 58.95%—the lowest among all comparing networks;
2) we unveil that VD-VGG with nine convolutional layers, and two FC layers offer the best performance. However, it requires a whole day to learn 10.4 M parameters with CPU;

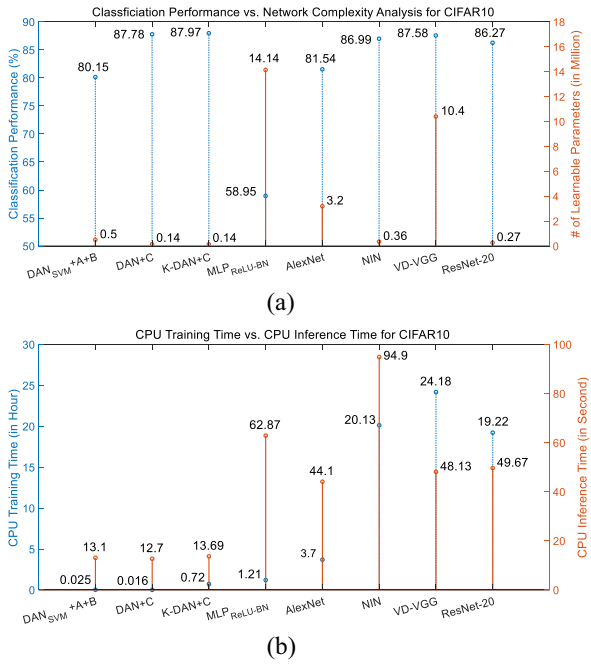

Fig. 7. (a) Classification performance versus network complexity and (b) CPU training time versus CPU interference time for DAN/K-DAN and BP-learned DNNs on CIFAR10.

3) in comparison with that of VD-VGG, each NIN and ResNet-20 learns a significantly reduced parameter set, specifically, only 0.36M for NIN and 0.27 for ResNet-20, while securing a considerable accuracy of 87%. Despite of that NIN and ResNet-20 expend 19 h at least in training as these two networks are generally more complicated. Therefore, GPU is indispensable to expedite the trivial FT task;
4) assuming that the PCA-compressed composite feature sets, that is, PCANet (A), and 2-FFCPCA (B) and the VD-VGG (C) features are obtainable for feature relearning. DAN/KDAN consume only less than 90 s to train 0.5M and 0.14M parameters, respectively;
5) aside from network complexity and CPU training time, it is worth mentioning that DAN/K-DAN necessitate negligibly low inference time.

Opposing to the BP networks, we reaffirm that learning the RR-driven DAN/K-DAN requires only minimum effort. If DAN/K-DAN is fine-tuned for each layer individually, we believe its performance especially in CIFAR10 will be improved.

## VI. Conclusion

S-DNN refers to a non-BP DNN assembled by means of aggregating nonlinear self-learnable building blocks, one after another, for a deep, feedforward construction. We outline a modularly trained S-DNN upon RR, dubbed DAN and its kernelized subsidiary (K-DAN) for pattern classification. We underline that:
1) training DAN/K-DAN is of noniterative and non-BP, but only minimal efforts since the one-shot analytic solution is learnable by only CPU, disregarding of training capacity;

2) opposing to other S-DNNs and the BP-trained networks learned only from the raw image pixels, DAN/K-DAN are demonstrated trainable on top of the pre-extracted baseline features (including the pretrained CNN features and the deliberately compressed features), and the structured features, that is, nonsignal or nonsequential data;
3) DAN/K-DAN operate more than a classifier, but triggering also feature relearning and feature compression;
4) we formulate a set of mathematical theories and proofs to reason the generalizability of DAN/K-DAN;
5) both DAN/K-DAN are revealed to be the most promising among other existing S-DNNs for a wide range of pattern classification tasks, for example, faces, handwritten digits, natural objects, and the structured features;
6) in the meantime, DAN/K-DAN stands out from other BP-optimized networks, in terms of network complexity, CPU training time, and CPU inference time.

For future exploration, it would be interesting if DAN/K-DAN also permits weight transferability like what BP-based DNNs do.

## References

[1] Y. LeCun, Y. Bengio, and G. Hinton, "Deep learning," *Nature*, vol. 521, no. 7553, pp. 436–444, May 2015.
[2] Y. Taigman, M. Yang, M. Ranzato, and L. Wolf, "DeepFace: Closing the gap to human-level performance in face verification," in *Proc. CVPR*, Columbus, OH, USA, 2014, pp. 1701–1708.
[3] Y. Sun, L. Ding, X. Wang, and X. Tang. (2015). *DeepID3: Face Recognition With Very Deep Neural Networks*. [Online]. Available: https://arXiv:1502.00873
[4] F. Schroff, D. Kalenichenko, and J. Philbin, "FaceNet: A unified embedding for face recognition and clustering," in *Proc. CVPR*, Boston, MA, USA, 2015, pp. 815–823.
[5] S. Hochreiter and J. Schmidhuber, "Long short-term memory," *Neural Comput.*, vol. 9, no. 8, pp. 1735–1780, Nov. 1997.
[6] H. Sak, A. Senior, and F. Beaufays. (2014). *Long Short-Term Memory Based Recurrent Neural Network Architectures*. [Online]. Available: https://arXiv:1402.1128
[7] D. H. Wolpert, "Stacked generalization," *Neural Netw.*, vol. 5, no. 2, pp. 241–259, 1992.
[8] L. Breiman, "Stacked regressions," *Mach. Learn.*, vol. 24, no. 1, pp. 49–64, 1996.
[9] M. Turk and A. Pentland, "Eigenfaces for recognition," *J. Cogn. Neurosci.*, vol. 3, no. 1, pp. 71–86, 1991.
[10] T.-H. Chan *et al.*, "PCANet: A simple deep learning baseline for image classification?" *IEEE Trans. Image Process.*, vol. 24, no. 12, pp. 5017–5032, Dec. 2015.
[11] J. Wu *et al.*, "Multilinear principal component analysis network for tensor object classification," *IEEE Access*, vol. 5, pp. 3322–3331, 2017.
[12] J. Huang and C. Yuan, "Weighted-PCANet for face recognition," in *Proc. Int. Conf. Neural Inf. Process.*, Nov. 2015, pp. 246–254.
[13] W.-L. Hao and Z. Zhang, "Incremental PCANet: A lifelong learning framework to achieve the plasticity of both feature and classifier constructions," in *Proc. Int. Conf. Brain Inspired Cogn. Syst.*, Nov. 2016, pp. 298–309.
[14] D. L. Swets and J. J. Weng, "Using discriminant eigenfeatures for image retrieval," *IEEE Trans. Pattern Anal. Mach. Intell.*, vol. 18, no. 8, pp. 831–836, Aug. 1996.
[15] Z. Lei, D. Yi, and S. Z. Li, "Learning stacked image descriptor for face recognition," *IEEE Trans. Circuits Syst. Video Technol.*, vol. 26, no. 9, pp. 1685–1696, Sep. 2016.
[16] G. E. Hinton, S. Osindero, and Y. W. Teh, "A fast learning algorithm for deep belief nets," *Neural Comput.*, vol. 18, no. 7, pp. 1527–1554, 2006.
[17] H. Lee, R. Grosse, R. Ranganath, and A. Y. Ng, "Convolutional deep belief networks for scalable unsupervised learning of hierarchical representation," in *Proc. ICML*, Montreal, QC, Canada, 2009, pp. 609–616.
[18] R. Salakhutdinov and H. Larochelle, "Efficient learning of deep Boltzmann machines," in *Proc. Mach. Learn. Res.*, vol. 9, 2010, pp. 693–700.

[19] P. Vincent, H. Larochelle, I. Lajoie, Y. Bengio, and P.-A. Manzagol, "Stacked denoising autoencoders: Learning useful representations in a deep network with a local denoising criterion," *J. Mach. Learn. Res.*, vol. 11, pp. 3371–3408, Jan. 2010.

[20] A. E. Hoerl and R. W. Kennard, "Ridge regression: Biased estimation for nonorthogonal problems," *Technometrics*, vol. 12, no. 1, pp. 55–67, 1970.

[21] M. A. Aizerman, E. A. Braverman, and L. I. Rozonoer, "Theoretical foundations of the potential function method in pattern recognition learning," *Autom. Remote Control*, vol. 25, pp. 821–837, Feb. 1964.

[22] A. Bordes, S. Ertekin, J. Weston, and L. Bottou, "Fast kernel classifiers with online and active learning," *J. Mach. Learn. Res.*, vol. 6, pp. 1579–1619, Sep. 2005.

[23] C. Cortes and V. Vapnik, "Support-vector networks," *Mach. Learn.*, vol. 20, no. 3, pp. 273–297, 1995.

[24] L. Deng, D. Yu, and J. Platt, "Scalable stacking and learning for building deep architectures," in *Proc. ICASSP*, Kyoto, Japan, 2012, pp. 2133–2136.

[25] L. Deng, G. Tur, X. He, and D. Hakkani-Tur, "Use of kernel deep convex networks and end-to-end learning for spoken language understanding," in *Proc. IEEE Workshop Spoken Lang. Technol.*, Miami, FL, USA, 2012, pp. 210–215.

[26] P.-S. Huang, L. Deng, M. Hasegawa-Johnson, and X. He, "Random features for kernel deep convex network," in *Proc. ICASSP*, Vancouver, BC, Canada, 2013, pp. 3143–3147.

[27] B. Hutchinson, L. Deng, and D. Yu, "Tensor deep stacking networks," *IEEE Trans. Pattern Anal. Mach. Intell.*, vol. 35, no. 8, pp. 1944–1957, Aug. 2013.

[28] G.-B. Huang, H. Zhou, X. Ding, and R. Zhang, "Extreme learning machine for regression and multiclass classification," *IEEE Trans. Syst., Man, Cybern. B, Cybern.*, vol. 42, no. 2, pp. 513–529, Apr. 2012.

[29] H. Zhou, G.-B. Huang, Z. Lin, H. Wang, and Y. C. Soh, "Stacked extreme learning machines," *IEEE Trans. Cybern.*, vol. 45, no. 9, pp. 2013–2025, Sep. 2015.

[30] J. Tang, C. Deng, and G.-B. Huang, "Extreme learning machine for multilayer perceptron," *IEEE Trans. Neural Netw. Learn. Syst.*, vol. 27, no. 4, pp. 809–821, Apr. 2016.

[31] L. Breiman, "Random forests," *Mach. Learn.*, vol. 45, no. 1, pp. 5–32, 2001.

[32] Z.-H. Zhou and J. Feng, "Deep forest: Towards an alternative to deep neural networks," in *Proc. IJCAI*, 2017, pp. 3553–3559.

[33] C. Hettinger. (Jun. 2017). *Forward Thinking: Building and Training Neural Networks One Layer at a Time*. [Online]. Available: https://arXiv:1706.02480

[34] L. V. Utkin, and M. A. Ryabinin. (Apr. 2017). *A Siamese Deep Forest*. [Online]. Available: https://arXiv:1704.08715

[35] X.-Z. Wang, T. Zhang, and R. Wang, "Noniterative deep learning: Incorporating restricted Boltzmann machine into multilayer random weight neural networks," *IEEE Trans Syst., Man, Cybern., Syst.*, to be published.

[36] P. J. Phillips, H. Moon, S. A. Rizvi, and P. J. Rauss, "The FERET evaluation methodology for face-recognition algorithms," *IEEE Trans. Pattern Anal. Mach. Intell.*, vol. 22, no. 10, pp. 1090–1101, Oct. 2000.

[37] Y. Lecun, L. Bottou, Y. Bengio, and P. Haffner, "Gradient-based learning applied to document recognition," *Proc. IEEE*, vol. 86, no. 11, pp. 2278–2324, Nov. 1998.

[38] A. Krizhevsky, "Learning multiple layers of features from tiny images," Dept. Comput. Sci., Univ. Toronto, Toronto, ON, Canada, Rep., 2009.

[39] F. F. Li, *Tiny ImageNet Visual Recognition Challenge*, Sandford Artif. Intell. Lab., Stanford Univ., Stanford, CA, USA, 2015. [Online]. Available: https://tiny-imagenet.herokuapp.com/

[40] O. Russakovsky *et al.*, "ImageNet large scale visual recognition challenge," *Int. J. Comput. Vis.*, vol. 115, no. 3, pp. 211–252, Dec. 2015.

[41] M. Lichman, *UCI Machine Learning Repositories*, Univ. California at Irvine, Irvine, CA, USA, 2013. [Online]. Available: http://archive.ics.uci.edu/ml

[42] J. Kannala and E. Rahtu, "BSIF: Binarized statistical image features," in *Proc. ICPR*, Tsukuba, Japan, 2012, pp. 1363–1366.

[43] C.-Y. Low, A. B.-J. Teoh, and C.-J. Ng, "Multi-fold Gabor, PCA, and ICA filter convolution descriptor for face recognition," *IEEE Trans. Circuits Syst. Video Technol.*, vol. 29, no. 1, pp. 115–129, Jan. 2019.

[44] C. Y. Low and A. B.-J. Teoh, "Stacking-based deep neural network: Deep analytic network on convolutional spectral histogram features," presented at the ICIP, 2017, pp. 1592–1596. [Online]. Available: https://arXiv:1703.01396

[45] L. Wan, M. Zeiler, S. Zhang, Y. LeCun, and R. Fergus, "Regularization of neural network using DropConnect," in *Proc. ICML*, 2013, pp. 1–9.

[46] K. He, X. Zhang, S. Ren, and J. Sun, "Spatial pyramid pooling in deep convolutional networks for visual recognition," in *Proc. ECCV*, 2014, pp. 346–361.

[47] M. D. Tissera and M. D. McDonnell, "Deep extreme learning machines: Supervised autoencoding architecture for classification," *Neurocomputing*, vol. 174, pp. 42–49, Jan. 2016.

[48] B. Graham. (May 2015). *Fractional Max-Pooling*. [Online]. Available: https://arXiv:1412.6071v4

[49] J. T. Springenberg, A. Dosovitskiy, T. Brox, and M. Riedmiller, "Striving for simplicity: The all convolutional net," in *Proc. ICLR Workshop*, May 2015. [Online]. Available: https://arXiv:1412.6806v3

[50] D. Mishkin and J. Matas, "All you need is a good init," in *Proc. ICLR*, May 2016. [Online]. Available: https://arXiv:1511.06422v7

[51] X. Glorot, A. Bordes, and Y. Bengio, "Deep sparse rectifier neural networks," in *Proc. Mach. Learn. Res.*, vol. 15, 2011, pp. 315–323.

[52] A. Vedaldi and K. Lenc, "MatConvNet: Convolutional neural networks for MATLAB," in *Proc. ACM MM*, Oct. 2015, pp. 689–692.

[53] C. M. Bishop, *Neural Networks for Pattern Recognition*, 3rd ed. New York, NY, USA: Oxford Univ. Press, 1995. [Online]. Available: https://dl.acm.org/citation.cfm?id=525960

[54] A. Krizhevsky, I. Sutskever, and G. E. Hinton, "ImageNet classification with deep convolutional neural networks," in *Proc. NIPS*, vol. 1, Dec. 2012, pp. 1097–1105.

[55] M. Lin, Q. Chen, and S. Yan, "Network in network," in *Proc. ICLR*, Apr. 2014. [Online]. Available: https://arXiv:1312.4400v3

[56] K. Simonyan and A. Zisserman, "Very deep convolutional networks for large-scale image recognition," in *Proc. ICLR*, May 2015. [Online]. Available: https://arXiv:1409.1556v6

[57] K. He, X. Zhang, S. Ren, and J. Sun. (2015). *Deep Residual Learning for Image Recognition*. [Online]. Available: https://arXiv:1512.03385

[58] J. Yosinski, J. Clune, Y. Bengio, and H. Lipson, "How transferable are features in deep neural networks," in *Proc. NIPS*, vol. 2. Montreal, QC, Canada, Dec. 2014, pp. 3320–3328.

[59] F. Chollet, "The limitations of deep learning," in *Deep Learning With Python*. New York, NY, USA: Manning, Nov. 2017, ch. 9. [Online]. Available: https://blog.keras.io/the-limitations-of-deep-learning.html

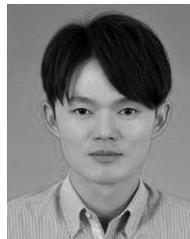

**Cheng-Yaw Low** received the Ph.D. degree in electrical and electronic engineering from Yonsei University, Seoul, South Korea, in 2018.

He is currently a Lecturer with the Faculty of Information Science and Technology, Multimedia University, Melaka, Malaysia. His current research interests include machine learning and pattern recognition.

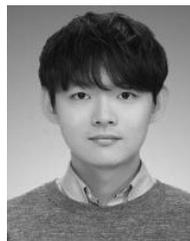

**Jaewoo Park** received the B.S. degree in mathematics from the University of Sydney, Sydney, NSW, Australia, in 2013 and the M.S. degree in mathematics from Yonsei University, Seoul, South Korea, in 2017, where he is currently pursuing the Ph.D. degree with the School of Electrical and Electronic Engineering.

His current research interests include unsupervised learning, specifics of which include disentangled representation, continuous learning, and generative model.

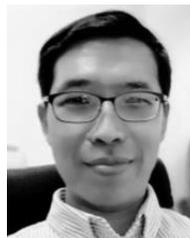

**Andrew Beng-Jin Teoh** (SM'12) received the Ph.D. degree in electrical, electronic, and system engineering from the National University of Malaysia, Bangi, Malaysia, in 2003.

He is a Professor with the School of Electrical and Electronic Engineering, Yonsei University, Seoul, South Korea. He has published over 300 international refereed journals, conference articles, and several book chapters in the areas of biometrics and machine learning.

Dr. Teoh served and is serving as a Guest Editor for the *IEEE Signal Processing Magazine* and an Associate Editor for the IEEE TRANSACTION ON INFORMATION FORENSIC AND SECURITY and the IEEE Biometrics Compendium.